\documentclass[runningheads]{llncs}
\usepackage{graphicx}
\usepackage{booktabs} 
\usepackage{amsfonts}
\usepackage{amsmath}
\usepackage[hidelinks]{hyperref}
\usepackage{url}
\usepackage[noend,algo2e,boxruled,linesnumbered]{algorithm2e}
\usepackage{multirow}
\newcommand{\abele}{{\scshape abele}}
\newcommand{\lime}{{\scshape lime}}
\newcommand{\lore}{{\scshape lore}}
\newcommand{\intg}{{\scshape intg}}
\newcommand{\grad}{{\scshape grad}}
\newcommand{\sal}{{\scshape sal}}
\newcommand{\occ}{{\scshape occ}}
\newcommand{\elrp}{{\scshape elrp}}
\newcommand{\mmd}{{\scshape mmd}}
\newcommand{\kmedoid}{{\scshape k-medoids}}
\newcommand\mydots{\makebox[1em][c]{.\hfil.\hfil.}}
\usepackage{pifont}
\newcommand{\cmark}{\ding{51}}%
\newcommand{\xmark}{\ding{55}}%

\begin{document}
\title{Black Box Explanation by Learning Image Exemplars in the Latent Feature Space\thanks{The final version of this work will appear in the proceedings of ECML-PKDD 2019. Please refer to the final version for future citations.}}

\titlerunning{Black Box Explanation by Learning Image Exemplars}

\author{
    Riccardo Guidotti\inst{1} \and
    Anna Monreale\inst{2} \and
    Stan Matwin \inst{3,4} \and
    Dino Pedreschi\inst{2}
}

%
%
\authorrunning{R. Guidotti et al.}
\tocauthor{Riccardo~Guidotti Anna~Monreale  Stan~Matwin Dino~Pedreschi}

\institute{
    ISTI-CNR, Pisa, Italy, \email{riccardo.guidotti@isti.cnr.it} \and
    University of Pisa, Italy, \email{\{name.surname\}@unipi.it} \and
    Dalhousie University,
    \email{stan@cs.dal.ca} \and Institute of Computer Scicne, Polish Academy of Sciences
}
\maketitle  
\begin{abstract}
We present an approach to explain the decisions of black box models for image classification. While using the black box to label images, our explanation method exploits the latent feature space learned through an adversarial autoencoder. The proposed method first generates exemplar images in the latent feature space and learns a decision tree classifier. Then, it selects and decodes exemplars respecting local decision rules. Finally, it visualizes them in a manner that shows to the user how the exemplars can be modified to either stay within their class, or to become counter-factuals by ``morphing'' into another class. Since we focus on black box decision systems for image classification, the explanation obtained from the exemplars also provides a saliency map highlighting the areas of the image that contribute to its classification, and areas of the image that push it into another class. We present the results of an experimental evaluation on three datasets and two black box models. Besides providing the most useful and interpretable explanations, we show that the proposed method outperforms existing explainers in terms of fidelity, relevance, coherence, and stability.

\keywords{Explainable AI, Adversarial Autoencoder, Image Exemplars.}
\end{abstract}
\section{Introduction}
\label{sec:introduction}
Automated decision systems based on machine learning techniques are widely used for classification, recognition and prediction tasks. These systems try to capture the relationships between the input instances and the target to be predicted. 
Input attributes can be of any type, as long as it is possible to find a convenient representation for them. 
For instance, we can represent images by matrices of pixels, or by a set of features that correspond to specific areas or patterns of the image.
Many automated decision systems are based on very accurate classifiers such as deep neural networks.
They are recognized to be ``black box" models because of their opaque, hidden internal structure, whose complexity makes their comprehension for humans very difficult~\cite{doshi2017towards}.
Thus, there is an increasing interest in the scientific community in deriving explanations able to describe the behavior of a black box~\cite{doshi2017towards,molnar2018interpretable,guidotti2018survey,escalante2018explainable}, or explainable by design approaches~\cite{li2018deep,kim2016examples}.
Moreover, the \textit{General Data Protection Regulation}\footnote{ \url{https://ec.europa.eu/justice/smedataprotect/}} has been approved in May 2018 by the European Parliament. 
This law gives to individuals the right to request ``...meaningful information of the logic involved'' when automated decision-making takes place with ``legal or similarly relevant effects'' on individuals.
Without a technology able to explain, in a manner easily understandable to a human, how a black box takes its decision, this right will remain only an utopia, or it will result in prohibiting the use of opaque, but highly effective machine learning methods in socially sensitive domains.

In this paper, we investigate the problem of black box explanation for image classification (Section~\ref{sec:problem}). 
Explaining the reasons for a certain decision can be particularly important.
For example, when dealing with medical images for diagnosing, how we can validate that a very accurate image classifier built to recognize cancer actually focuses on the malign areas and not on the background for taking the decisions?

In the literature (Section~\ref{sec:related}), the problem is addressed by producing explanations through different approaches.
On the one hand, gradient and perturbation-based attribution methods~\cite{simonyan2013deep,shrikumar2016not} reveal saliency maps highlighting the parts of the image that most contribute to its classification.
However, these methods are \textit{model specific} and can be employed only to explain specific deep neural networks.
On the other hand, \textit{model agnostic} approaches can explain, yet through a saliency map, the outcome of any black box~\cite{ribeiro2016should,guidotti2019investigating}.
Agnostic methods may generate a local neighborhood of the instance to explain and mime the behavior of the black box using an interpretable classifier.
However, these methods exhibit drawbacks that may negatively impact the reliability of the explanations.
First, they do not take into account existing relationships between features (or pixels) during the neighborhood generation.
Second, the neighborhood generation does not produce ``meaningful'' images since, e.g., some areas of the image to explain in~\cite{ribeiro2016should} are obscured, while in~\cite{guidotti2019investigating} they are replaced with pixels of other images. 
Finally, transparent-by-design approaches produce prototypes from which it should be clear to the user why a certain decision is taken by the model~\cite{kim2016examples,li2018deep}.
Nevertheless, these approaches cannot be used to explain a trained black box, but the transparent model has to be directly adopted as a classifier, possibly with limitations on the accuracy achieved.

We propose \abele, an Adversarial Black box Explainer generating Latent Exemplars (Section~\ref{sec:method}).
\abele\ is a local, model-agnostic explanation method able to overcome the existing limitations of the local approaches by exploiting the latent feature space, learned through an adversarial autoencoder~\cite{makhzani2015adversarial} (Section~\ref{sec:AAE}), for the neighborhood generation process.
Given an image classified by a given black box model, \abele\ provides an explanation for the reasons of the proposed classification. 
The explanation consists of two parts:
\textit{(i)} a set of \textit{exemplars} and \textit{counter-exemplars} images illustrating, respectively, instances classified with the same label and with a different label than the instance to explain, which may be visually analyzed to understand the reasons for the classification, and \textit{(ii)}
a \textit{saliency map} highlighting the areas of the image to explain that contribute to its classification, and areas of the image that push it towards another label. 

We present a deep experimentation (Section~\ref{sec:experiments}) on three datasets of images, i.e., \texttt{mnist}, \texttt{fashion} and \texttt{cifar10}, and two black box models.
We empirically prove that \abele\ overtakes state of the art methods based on saliency maps or on prototype selection by providing relevant, coherent, stable and faithful explanations.
Finally, we summarize our contribution, its limitations, and future research directions (Section~\ref{sec:conclusion}).

\section{Related Work}
\label{sec:related}
Research on black box explanation methods has recently received much attention~\cite{doshi2017towards,molnar2018interpretable,guidotti2018survey,escalante2018explainable}.
These methods can be characterized as model-specific \textit{vs} model-agnostic, and local \textit{vs} global. 
The proposed explanation method \abele\ is the next step in the line of research on local, model-agnostic methods originated with~\cite{ribeiro2016should} and extended in different directions by~\cite{frosst2018distilling} and by~\cite{guidotti2018local,guidotti2019investigating,panigutti2019explaining}.

In image classification, typical explanations are the \textit{saliency maps}, i.e., images that show each pixel's positive (or negative) contribution to the black box outcome.
Saliency maps are efficiently built by gradient~\cite{simonyan2013deep,shrikumar2016not,sundararajan2017axiomatic,bach2015pixel} and perturbation-based~\cite{zeiler2014visualizing,fong2017interpretable} attribution methods by finding, through backpropagation and differences on the neuron activation, the pixels of the image that maximize an approximation of a linear model of the black box classification outcome. Unfortunately, these approaches are specifically designed for deep neural networks.
They cannot be employed for explaining other image classifiers, like tree ensembles or hybrid image classification processes~\cite{guidotti2018survey}.
Model-agnostic explainers, such as \lime~\cite{ribeiro2016should} and similar~\cite{guidotti2019investigating} can be employed to explain the classification of any image classifier.
They are based on the generation of a local neighborhood around the image to explain, and on the training of an interpretable classifier on this neighborhood.
Unlike the global distillation methods~\cite{hinton2015distilling}, they do not consider (often non-linear) relationships between features (e.g. pixel proximity), and thus, their neighborhoods do not contain ``meaningful'' images.

Our proposed method \abele\ overcomes the limitations of both saliency-based and local model-agnostic explainers by using AAEs, local distillation, and exemplars. 
As \abele\ includes and extends \lore~\cite{guidotti2018local}, an innovation w.r.t. state of the art explainers for image classifiers is the usage of counter-factuals. 
Counter-factuals are generated from ``positive'' instances by a minimal perturbation that pushes them to be classified with a different label~\cite{van2018contrastive}. 
In line with this approach, \abele\ generates counter-factual rules in the latent feature space and exploits them to derive counter-exemplars in the original feature space.

As the explanations returned by \abele\ are based on exemplars, we need to clarify the relationship between \textit{exemplars} and \textit{prototypes}. 
Both are used as a foundation of representation of a category, or a concept~\cite{frixione2012prototypes}. 
In the prototype view, a concept is the representation of a specific instance of this concept. 
In the exemplar view, the concept is represented by means of a set of typical examples, or exemplars. 
\abele\ uses exemplars to represent a concept. 
In recent works~\cite{li2018deep,chen2018looks}, image prototypes are used as the foundation of the concept for interpretability~\cite{bien2011prototype}. 
In~\cite{li2018deep}, an explainable by design method,
similarly to \abele, generates prototypes in the latent feature space learned with an autoencoder.
However, it is not aimed at explaining a trained black box model. 
In~\cite{chen2018looks} a convolutional neural network is adopted to provide features from which the prototypes are selected. 
\abele\ differs from these approaches because is model agnostic and the \textit{adversarial} component ensures the similarity of feature and class distributions.

\section{Problem Formulation}
\label{sec:problem}
In this paper we address the \textit{black box outcome explanation problem}~\cite{guidotti2018survey}.
Given a black box model $b$ and an instance $x$ classified by $b$, i.e., $b(x) = y$, our aim is to provide an explanation $e$ for the decision $b(x)=y$.
More formally:
\begin{definition}\label{def:problem}
{\em
Let $b$ be a black box, and $x$ an instance whose decision $b(x)$ has to be explained. 
The \emph{black box outcome explanation problem} consists in finding an explanation $e \in E$ belonging to a human-interpretable domain $E$.
}
\end{definition}

We focus on the black box outcome explanation problem for image classification, where the instance $x$ is an image mapped by $b$ to a class label $y$.
In the following, we use the notation $b(X) = Y$ as a shorthand for $\{b(x) \;|\; x \in X\} = Y$.
We denote by $b$ a black box image classifier, whose internals are either unknown to the observer or they are known but uninterpretable by humans.
Examples are neural networks and ensemble classifiers.
We assume that a black box $b$ is a function that can be queried at will.

We tackle the above problem by deriving an explanation from the understanding of the behavior of the black box in the local neighborhood of the instance to explain~\cite{guidotti2018survey}.
To overcome the state of the art limitations, we exploit adversarial autoencoders~\cite{makhzani2015adversarial} for generating, encoding and decoding the local neighborhood. 
\begin{figure}[t]
    \centering
    \includegraphics[trim = 0mm 0mm 0mm 0mm, clip,width=\linewidth]{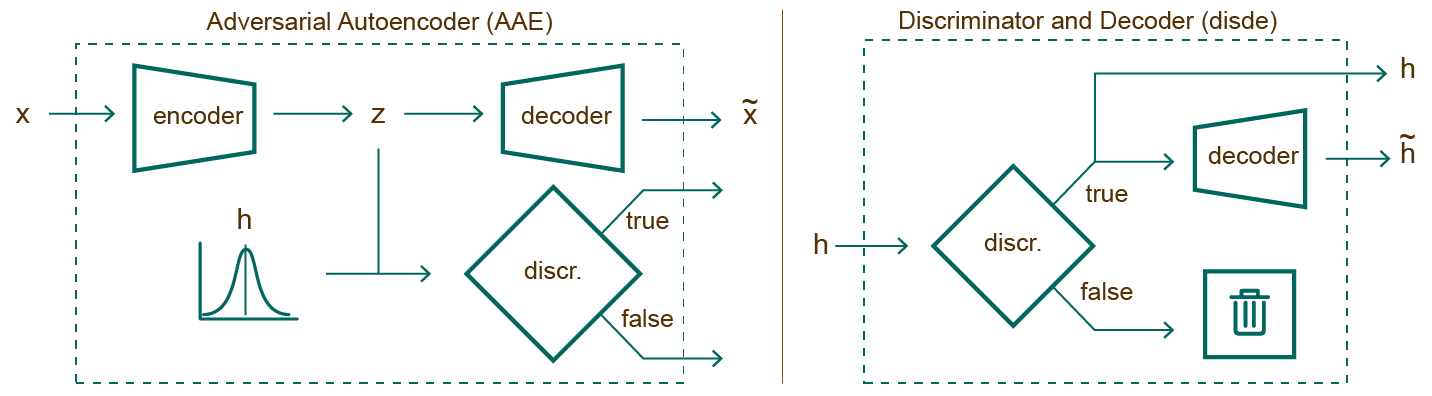}
    \caption{\textit{Left}: Adversarial Autoencoder architecture: the $\mathit{encoder}$ turns the image $x$ into its latent representation $z$, the $\mathit{decoder}$ re-builds an approximation $\widetilde{x}$ of $x$ from $z$, and the $\mathit{discriminator}$ identifies if a randomly generated latent instance $h$ can be considered valid or not.
    \textit{Right}: Discriminator and Decoder ($\mathit{disde}$) module: input is a randomly generated latent instance $h$ and, if it is considered valid by the $\mathit{discriminator}$, it returns it together with its decompressed version $\widetilde{h}$.}
    \label{fig:AAE}
\end{figure}

\section{Adversarial Autoencoders}
\label{sec:AAE}
An important issue arising in the use of synthetic instances generated when developing black box explanations is the question of maintaining the identity of the distribution of the examples that are generated with the prior distribution of the original examples. 
We approach this issue by using an Adversarial Autoencoder (AAE)~\cite{makhzani2015adversarial}, which combines a Generative Adversarial Network (GAN)~\cite{goodfellow2014generative} with the autoencoder representation learning algorithm.
Another reason for the use of AAE is that, as demonstrated in~\cite{sun2019enhancing}, the use of autoencoders enhances the robustness of deep neural network classifiers more against malicious examples.

AAEs are probabilistic autoencoders that aim at generating new random items that are highly similar to the training data. 
They are regularized by matching the aggregated posterior distribution of the latent representation of the input data to an arbitrary prior distribution. 
The AAE architecture (Fig.~\ref{fig:AAE}-left) includes an $\mathit{encoder}: \mathbb{R}^n {\rightarrow} \mathbb{R}^k$, a $\mathit{decoder}: \mathbb{R}^k {\rightarrow} \mathbb{R}^n$ and a $\mathit{discriminator}: \mathbb{R}^k {\rightarrow} [0, 1]$ where $n$ is the number of pixels in an image and $k$ is the number of latent features.
Let $x$ be an instance of the training data, we name $z$ the corresponding latent data representation obtained by the $\mathit{encoder}$.
We can describe the AAE with the following distributions~\cite{makhzani2015adversarial}: the prior distribution $p(z)$ to be imposed on $z$, the data distribution $p_d(x)$, the model distribution $p(x)$, and the encoding and decoding distributions $q(z|x)$ and $p(x|z)$, respectively.
The encoding function $q(z|x)$ defines an aggregated posterior distribution of $q(z)$ on the latent feature space: $q(z) {=} \int_x q(z|x) p_d(x) dx$.
The AAE guarantees that the aggregated posterior distribution $q(z)$ matches the prior distribution $p(z)$, through the latent instances and by minimizing the reconstruction error. 
The AAE generator corresponds to the encoder $q(z|x)$ and ensures that the aggregated posterior distribution can confuse the $\mathit{discriminator}$ in deciding if the latent instance $z$ comes from the true distribution $p(z)$.

The AAE learning involves two phases: the \textit{reconstruction} aimed at training the $\mathit{encoder}$ and $\mathit{decoder}$ to minimize the reconstruction loss; the \textit{regularization} aimed at training the $\mathit{discriminator}$ using training data and encoded values.
After the learning, the decoder defines a generative model mapping $p(z)$ to $p_d(x)$.

\begin{figure}[t]
    \centering
    \includegraphics[trim = 0mm 0mm 0mm 0mm, clip,width=\linewidth]{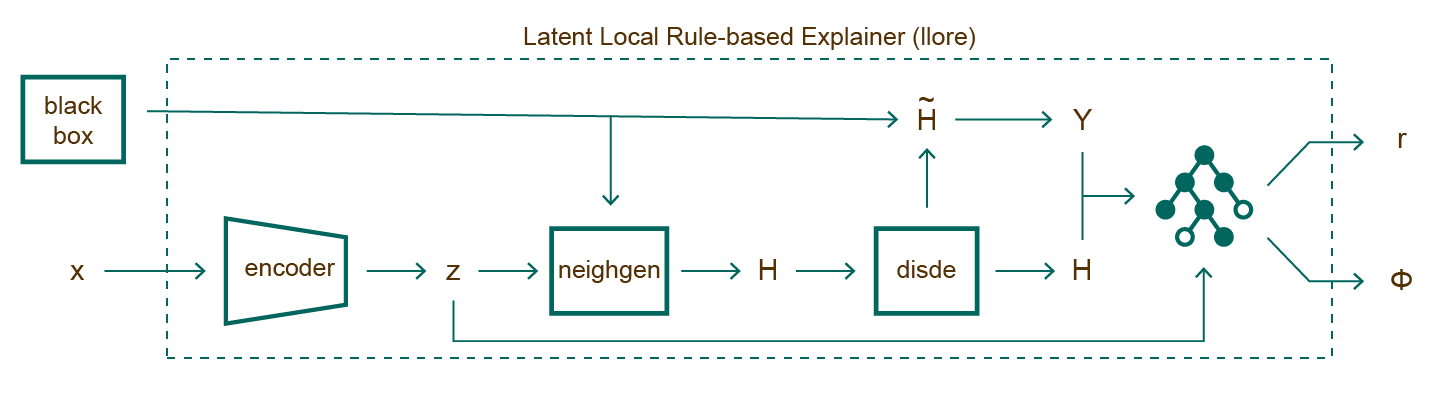}
    \caption{Latent Local Rules Extractor. It takes as input the image $x$ to explain and the black box $b$. With the $\mathit{encoder}$ trained by the AAE, it turns $x$ into its latent representation $z$. Then, the $\mathit{neighgen}$ module uses $z$ and $b$ to generate the latent local neighborhood $H$. The valid instances are decoded in $\widetilde{H}$ by the $\mathit{disde}$ module. Images in $\widetilde{H}$ are labeled with the black box $Y = b(\widetilde{H})$. $H$ and $Y$ are used to learn a decision tree classifier. At last, a decision rule $r$ and the counter-factual rules $\Phi$ for $z$ are returned.}
    
    \label{fig:learning}
\end{figure}

\section{Adversarial Black Box Explainer 
}
\label{sec:method}
\abele\ (Adversarial Black box Explainer generating Latent Exemplars) is a local model agnostic explainer for image classifiers solving the outcome explanation problem.
Given an image $x$ to explain and a black box $b$, the explanation provided by \abele\ is composed of \textit{(i)} a set of \textit{exemplars} and \textit{counter-exemplars}, \textit{(ii)} a \textit{saliency map}.
Exemplars and counter-exemplars shows instances classified with the same and with a different outcome than $x$.
They can be visually analyzed to understand the reasons for the decision.
The saliency map highlights the areas of $x$ that contribute to its classification and areas that push it into another class.
The explanation process involves the following steps.
First, \abele\ generates a neighborhood in the latent feature space exploiting the AAE (Sec.~\ref{sec:AAE}).
Then, it learns a decision tree on that latent neighborhood providing local decision and counter-factual rules.
Finally, \abele\ selects and decodes exemplars and counter-exemplars satisfying these rules and extracts from them a saliency map.

\subsection{Encoding}
The image $x {\in} \mathbb{R}^n$ to be explained is passed as input to the AAE where the $\mathit{encoder}$ returns the latent representation $z \in \mathbb{R}^k$ using $k$ latent features with ${k \ll n}$. The number $k$ is kept low by construction avoiding high dimensionality problems.

\subsection{Neighborhood Generation}
\abele\ generates a set $H$ of $N$ instances in the latent feature space, with characteristics close to those of $z$. 
Since the goal is to learn a predictor on $H$ able to simulate the local behavior of $b$, the neighborhood includes instances with both decisions, i.e., $H = H_{=} \cup H_{\neq}$ where instances $h \in H_=$ are such that $b(\widetilde{h}) = b(x)$, and $h \in H_{\neq}$ are such that $b(\widetilde{h}) \neq b(x)$. 
We name $\widetilde{h} \in \mathbb{R}^n$ the decoded version of an instance $h \in \mathbb{R}^k$ in the latent feature space.
The neighborhood generation of $H$ ($\mathit{neighgen}$ module in Fig.~\ref{fig:learning}) may be accomplished using different strategies ranging from pure random strategy using a given distribution to a genetic approach 
maximizing a fitness function~\cite{guidotti2018local}. 
In our experiments we adopt the last strategy.
After the generation process, for any instance $h \in H$, \abele\ exploits the $\mathit{disde}$ module (Fig.~\ref{fig:AAE}-right
) for both checking the validity of $h$ by querying the $\mathit{discriminator}$\footnote{In the experiments we use for the $\mathit{discriminator}$ the default validity threshold $0.5$ to distinguish between real and fake exemplars. This value can be increased to admit only more reliable exemplars, or decreased to speed-up the generation process.} and decoding it into $\widetilde{h}$.
Then, \abele\ queries the black box $b$ with $\widetilde{h}$ to get the class $y$, i.e.,~$b(\widetilde{h})=y$.

\subsection{Local Classifier Rule Extraction}
Given the local neighborhood $H$, \abele\ builds a decision tree classifier $c$ trained on the instances $H$ labeled with the black box decision $b(\widetilde{H})$. 
Such a predictor is intended to locally mimic the behavior of $b$ in the neighborhood $H$. 
The decision tree extracts the decision rule $r$ and counter-factual rules $\Phi$ enabling the generation of \textit{exemplars} and \textit{counter-exemplars}.
\abele\ considers decision tree classifiers because: \emph{(i)} decision rules can naturally be derived from a root-leaf path in a decision tree; and, \emph{(ii)} counter-factual rules can be extracted by symbolic reasoning over a decision tree.
The premise $p$ of a decision rule $r {=} p {\rightarrow} y$ is the conjunction of the splitting conditions in the nodes of the path from the root to the leaf that is satisfied by the latent representation $z$ of the instance to explain $x$, and setting $y {=} c(z)$.
For the counter-factual rules $\Phi$, \abele\ 
selects the closest rules in terms of splitting conditions leading to a label $\hat{y}$ different from $y$, i.e., the rules $\{q {\rightarrow} \hat{y}\}$ such that $q$ is the conjunction of splitting conditions for a path from the root to the leaf labeling an instance $h_c$ with $c(h_c) {=} \hat{y}$ and minimizing the number of splitting conditions falsified w.r.t. the premise $p$ of the rule $r$. 
Fig.~\ref{fig:learning} shows the process that, starting from the image to be explained, leads to the decision tree learning, and to the extraction of the decision and counter-factual rules. 
We name this module $\mathit{llore}$, as a variant of \lore~\cite{guidotti2018local} operating in the latent feature space. 
\begin{figure}[t]
    \centering
    \includegraphics[trim = 0mm 0mm 0mm 0mm, clip,width=\linewidth]{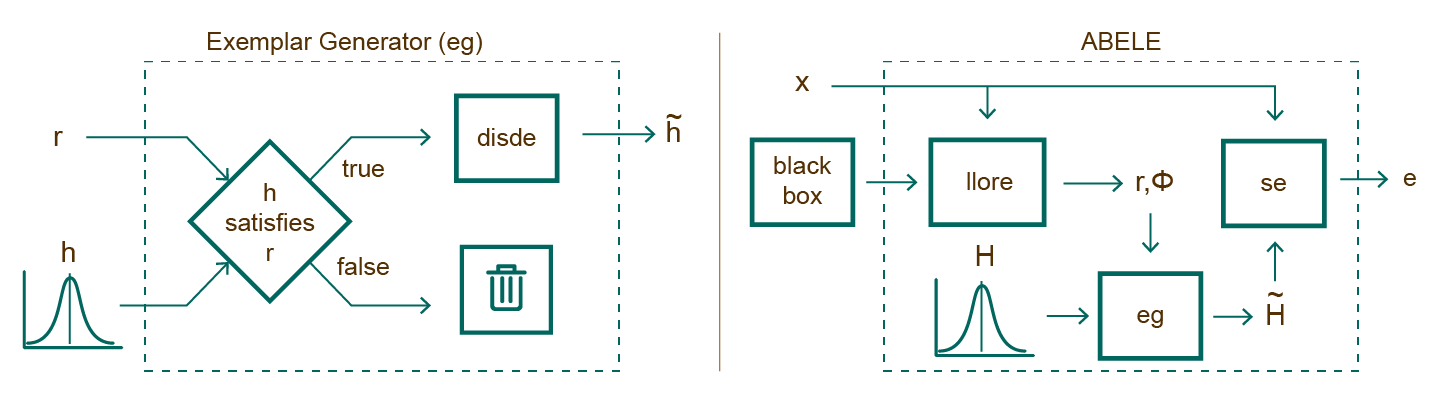}
    \caption{\textit{Left}: (Counter-)Exemplar Generator: it takes a decision rule $r$ and a randomly generated latent instance $h$, checks if $h$ satisfies $r$ and applies the $\mathit{disde}$ module (Fig.\ref{fig:AAE}-right) to decode it.
    \textit{Right}: \abele\ architecture. It takes as input the image $x$ for which we require an explanation and the black box $b$. It extracts the decision rule $r$ and the counter-factual rules $\Phi$ with the $\mathit{llore}$ module. Then, it generates a set of latent instances $H$ which are used as input with $r$ and $\Phi$ for the $\mathit{eg}$ module (Fig. \ref{fig:abele}-left) to generate exemplars and counter-exemplars $\widetilde{H}$. Finally, $x$ and $\widetilde{H}$ are used by the $\mathit{se}$ module for calculating the saliency maps and returning the final explanation $e$.}
    \label{fig:abele}
\end{figure}

\subsection{Explanation Extraction}
Often, e.g.~in medical or managerial decision making, people explain their decisions by pointing to exemplars with the same (or different) decision outcome~\cite{frixione2012prototypes,chen2018looks}. 
We follow this approach and we model the explanation of an image $x$ returned by \abele\ as a triple ${e=\langle \widetilde{H}_e, \widetilde{H}_c, s \rangle}$ composed by \textit{exemplars} $\widetilde{H}_e$, \textit{counter-exemplars} $\widetilde{H}_c$ and a \textit{saliency map} $s$.
Exemplars and counter-exemplars are images representing instances similar to $x$, leading to an outcome equal to or different from $b(x)$. 
Exemplars and counter-exemplars are generated by \abele\ exploiting the $\mathit{eg}$ module (Fig.~\ref{fig:abele}-left
). 
It first generates a set of latent instances $H$ satisfying the decision rule $r$ (or a set of counter-factual rules $\Phi$), as shown in Fig.~\ref{fig:learning}.
Then, it validates and decodes them into exemplars $\widetilde{H}_e$ (or counter-exemplars $\widetilde{H}_c$) using the $\mathit{disde}$ module.
The saliency map $s$ highlights areas of $x$ that contribute to its outcome and areas that push it into another class.
The map is obtained by the saliency extractor $\mathit{se}$ module (Fig.~\ref{fig:abele}-right) that first computes the pixel-to-pixel-difference between $x$ and each exemplar in the set  $\widetilde{H}_e$, and then, it assigns to each pixel of the saliency map $s$ the median value of all differences calculated for that pixel.
Thus, formally for each pixel $i$ of the saliency map $s$ we have: 
$s[i] = \mathit{median}_{\forall \widetilde{h}_e \in \widetilde{H}_e}(x[i] - \widetilde{h}_e[i]).$

In summary, \abele\ (Fig. \ref{fig:abele}-right), takes as input the instance to explain $x$ and a black box $b$, and returns an explanation $e$ according to the following steps.
First, it adopts $\mathit{llore}$ \cite{guidotti2018local} to extract the decision rule $r$ and the counterfactual rules $\Phi$.
These rules, together with a set of latent random instances $H$ are the input of the $\mathit{eg}$ module returning \textit{exemplars} and \textit{counter-exemplars}. Lastly, the $\mathit{se}$ module extracts the \textit{saliency map} starting from the image $x$ and its exemplars.

\section{Experiments}
\label{sec:experiments}
\begin{table}[t]
\begin{minipage}{.6\linewidth}
\footnotesize
\setlength{\tabcolsep}{1.2mm}
\centering
\begin{tabular}{|c|cccc|cc|}
\hline
dataset & resolution & rgb & train & test & RF & DNN \\
\hline
\texttt{mnist} & $28\times28$ & \xmark & $60k$ & $10k$ & $.9692$ & $.9922$ \\
\texttt{fashion} & $28\times28$ & \xmark & $60k$ & $10k$ & $.8654$ & $.9207$ \\
\texttt{cifar10} & $32\times32$ & \cmark & $50k$ & $10k$ & $.4606$ & $.9216$ \\
\hline
\end{tabular}
\caption{Datasets resolution, type of color, train and test dimensions, and black box model accuracy.}
\label{tab:datasetbb}
\end{minipage}
\hspace{2mm}
\begin{minipage}{.34\linewidth}
\footnotesize
\setlength{\tabcolsep}{1mm}
\centering
\begin{tabular}{|c|cc|}
\hline
dataset & train & test \\
\hline
\texttt{mnist} & $39.80$ & $43.64$ \\
\texttt{fashion} & $27.41$ & $30.15$  \\
\texttt{cifar10} & $20.26$ & $45.12$ \\
\hline
\end{tabular}
\caption{AAEs reconstruction error in terms of RMSE.}
\label{tab:autoencoders}
\end{minipage}
\end{table}
We experimented with the proposed approach on three open source datasets\footnote{Dataset:
\url{http://yann.lecun.com/exdb/mnist/},
\url{https://www.cs.toronto.edu/~kriz/cifar.html},
\url{https://www.kaggle.com/zalando-research/}.
} (details in Table~\ref{tab:datasetbb}): the \texttt{mnist} dataset of handwritten digit grayscale images, the \texttt{fashion} mnist dataset is a collection of Zalando's article grayscale images (e.g. shirt, shoes, bag, etc.), and the \texttt{cifar10} dataset of colored images of airplanes, cars, birds, cats, etc. 
Each dataset has ten different labels.

We trained and explained away the following black box classifiers.
Random Forest~\cite{breiman2001random} (RF) as implemented by the \textit{scikit-learn} Python library, and Deep Neural Networks (DNN) implemented with the \texttt{keras} library\footnote{Black box: 
\url{https://scikit-learn.org/}, \url{https://keras.io/examples/}.
}. 
For \texttt{mnist} and \texttt{fashion} we used a three-layer CNN, while for \texttt{cifar10} we used the \textit{ResNet20 v1} network described in~\cite{he2016deep}.
Classification performance are reported in Table~\ref{tab:datasetbb}.

For \texttt{mnist} and \texttt{fashion} we trained AAEs with sequential three-layer encoder, decoder and discriminator.
For \texttt{cifar10} we adopted a four-layer CNN for the encoder and the decoder, and a sequential discriminator.
We used 80\% of the test sets for training the adversarial autoencoders\footnote{The encoding distribution of AAE is defined as a Gaussian distribution whose mean and variance is predicted by the encoder itself \cite{makhzani2015adversarial}. 
We adopted the following number of latent features $k$ for the various datasets: \texttt{mnist} $k{=}4$, \texttt{fashion} $k{=}8$, \texttt{cifar10} $k{=}16$.}.
In Table~\ref{tab:autoencoders} we report the 
reconstruction error of the AAE in terms of \textit{Root Mean Square Error} (RMSE) between the original and reconstructed images.
We employed the remaining 20\% for evaluating the quality of the explanations.

We compare \abele\ against \lime\ and a set of saliency-based explainers collected in the \texttt{DeepExplain} package\footnote{Github code links: 
\url{https://github.com/riccotti/ABELE},
\url{https://github.com/marcotcr/lime},
\url{https://github.com/marcoancona/DeepExplain}
.}: Saliency (\sal)~\cite{simonyan2013deep}, GradInput (\grad)~\cite{shrikumar2016not}, IntGrad (\intg)~\cite{sundararajan2017axiomatic}, $\varepsilon$-lrp (\elrp)~\cite{bach2015pixel}, and Occlusion (\occ)~\cite{zeiler2014visualizing}.
We refer to the set of tested \texttt{DeepExplain} methods as {\scshape dex}.
We also compare the exemplars and counter-exemplars generated by \abele\ against the prototypes and criticisms\footnote{Criticisms are images not well-explained by prototypes with a regularized
kernel function~\cite{kim2016examples}.} selected by the \mmd\ and \kmedoid~\cite{kim2016examples}.
\mmd\ exploits the maximum mean discrepancy and a kernel function for selecting the best prototypes and criticisms.

\begin{figure}[t]
    \centering
    \begin{minipage}{.48\textwidth}
    \includegraphics[trim = 10mm 10mm 2mm 2mm, clip,width=0.12\linewidth]{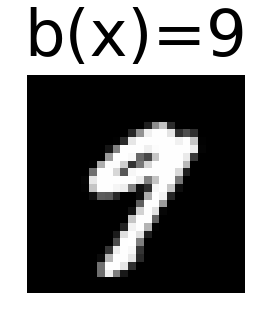}
    \includegraphics[trim = 10mm 10mm 2mm 2mm, clip,width=0.12\linewidth]{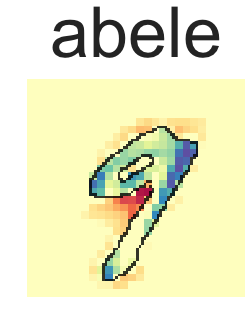}
    \includegraphics[trim = 10mm 10mm 2mm 2mm, clip,width=0.12\linewidth]{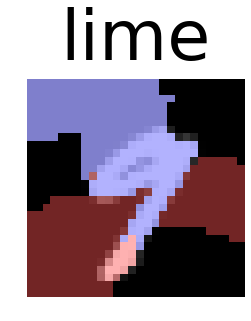}%
    \includegraphics[trim = 10mm 10mm 2mm 2mm, clip,width=0.12\linewidth]{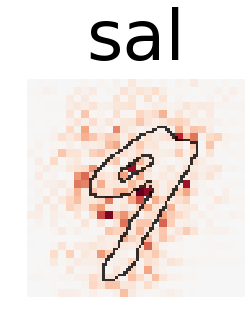}%
    \includegraphics[trim = 10mm 10mm 2mm 2mm, clip,width=0.12\linewidth]{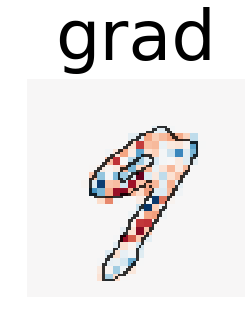}%
    \includegraphics[trim = 10mm 10mm 2mm 2mm, clip,width=0.12\linewidth]{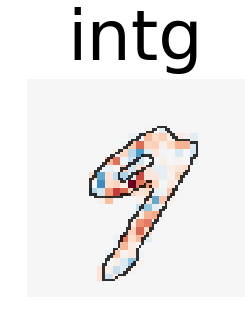}%
    \includegraphics[trim = 10mm 10mm 2mm 2mm, clip,width=0.12\linewidth]{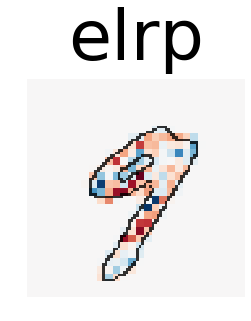}\\
    \includegraphics[trim = 10mm 10mm 2mm 2mm, clip,width=0.12\linewidth]{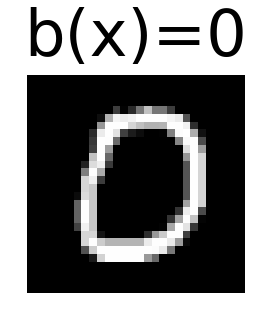}
    \includegraphics[trim = 10mm 10mm 2mm 2mm, clip,width=0.12\linewidth]{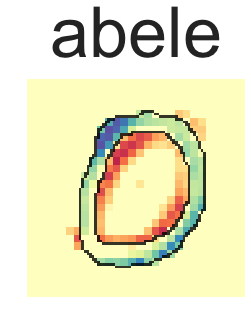}
    \includegraphics[trim = 10mm 10mm 2mm 2mm, clip,width=0.12\linewidth]{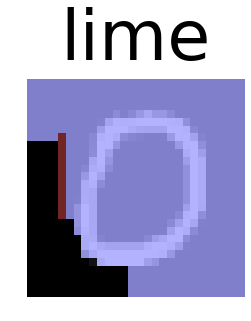}%
    \includegraphics[trim = 10mm 10mm 2mm 2mm, clip,width=0.12\linewidth]{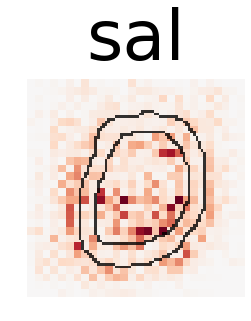}%
    \includegraphics[trim = 10mm 10mm 2mm 2mm, clip,width=0.12\linewidth]{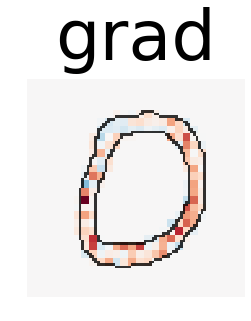}%
    \includegraphics[trim = 10mm 10mm 2mm 2mm, clip,width=0.12\linewidth]{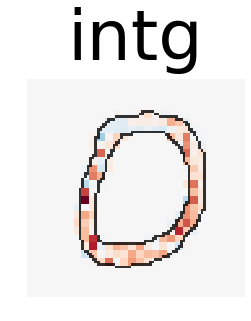}%
    \includegraphics[trim = 10mm 10mm 2mm 2mm, clip,width=0.12\linewidth]{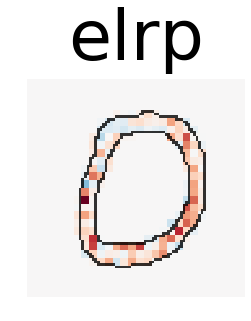}\\
    \includegraphics[trim = 10mm 10mm 2mm 2mm, clip,width=0.12\linewidth]{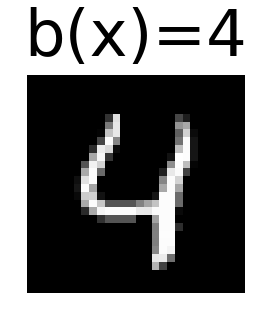}
    \includegraphics[trim = 10mm 10mm 2mm 2mm, clip,width=0.12\linewidth]{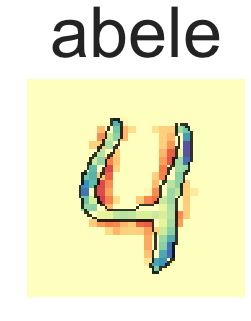}
    \includegraphics[trim = 10mm 10mm 2mm 2mm, clip,width=0.12\linewidth]{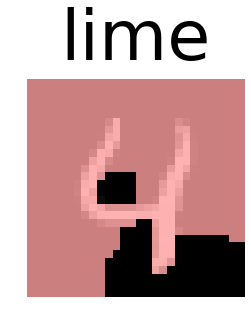}%
    \includegraphics[trim = 10mm 10mm 2mm 2mm, clip,width=0.12\linewidth]{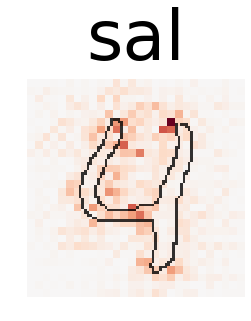}%
    \includegraphics[trim = 10mm 10mm 2mm 2mm, clip,width=0.12\linewidth]{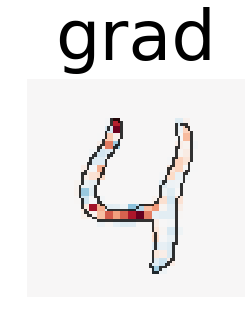}%
    \includegraphics[trim = 10mm 10mm 2mm 2mm, clip,width=0.12\linewidth]{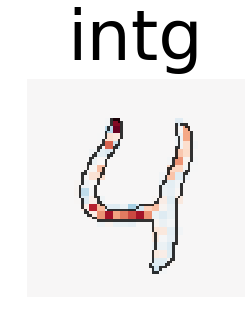}%
    \includegraphics[trim = 10mm 10mm 2mm 2mm, clip,width=0.12\linewidth]{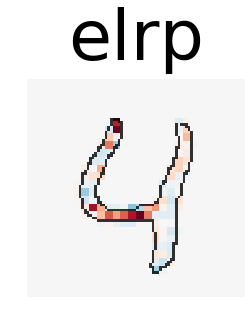}%
    \caption{Explain by saliency map \texttt{mnist}.}
    \label{fig:explanations_mnist}
    \end{minipage}%
\hspace{1mm}
    \begin{minipage}{.48\textwidth}
    \centering
    \includegraphics[trim = 10mm 10mm 2mm 2mm, clip,width=0.12\linewidth]{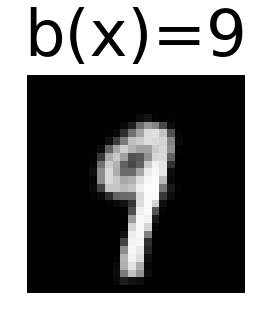}
    \includegraphics[trim = 10mm 10mm 2mm 2mm, clip,width=0.12\linewidth]{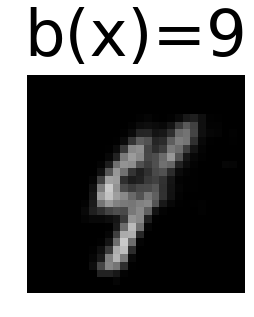}
    \includegraphics[trim = 10mm 10mm 2mm 2mm, clip,width=0.12\linewidth]{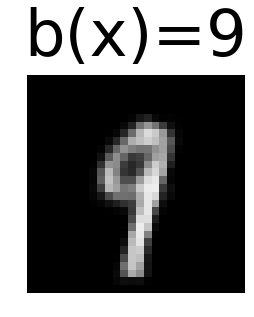}
    \hspace{1mm}
    \includegraphics[trim = 10mm 10mm 2mm 2mm, clip,width=0.12\linewidth]{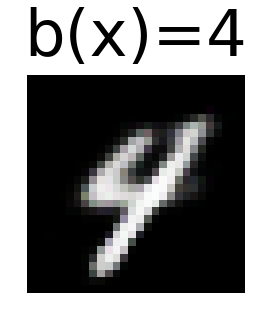}
    \includegraphics[trim = 10mm 10mm 2mm 2mm, clip,width=0.12\linewidth]{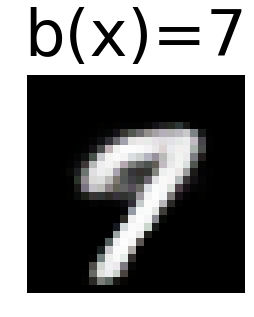}\\
    \includegraphics[trim = 10mm 10mm 2mm 2mm, clip,width=0.12\linewidth]{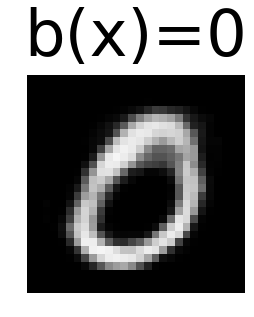}
    \includegraphics[trim = 10mm 10mm 2mm 2mm, clip,width=0.12\linewidth]{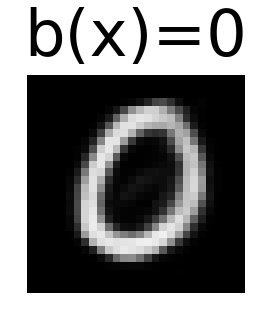}
    \includegraphics[trim = 10mm 10mm 2mm 2mm, clip,width=0.12\linewidth]{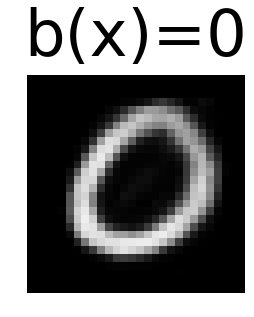}
    \hspace{1mm}
    \includegraphics[trim = 10mm 10mm 2mm 2mm, clip,width=0.12\linewidth]{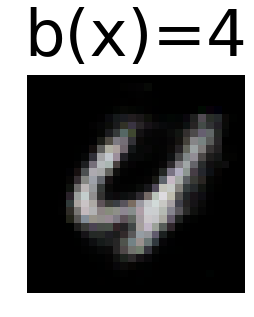}
    \includegraphics[trim = 10mm 10mm 2mm 2mm, clip,width=0.12\linewidth]{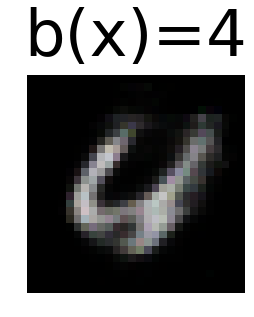}\\
    \includegraphics[trim = 10mm 10mm 2mm 2mm, clip,width=0.12\linewidth]{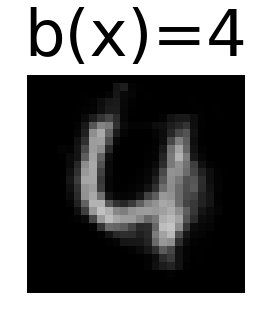}
    \includegraphics[trim = 10mm 10mm 2mm 2mm, clip,width=0.12\linewidth]{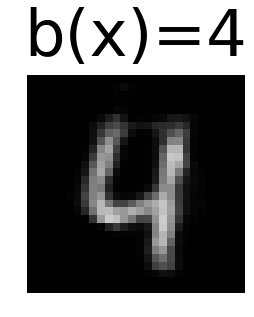}
    \includegraphics[trim = 10mm 10mm 2mm 2mm, clip,width=0.12\linewidth]{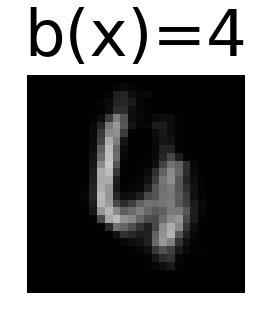}
    \hspace{1mm}
    \includegraphics[trim = 10mm 10mm 2mm 2mm, clip,width=0.12\linewidth]{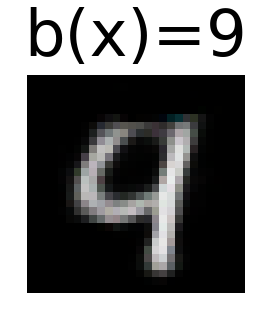}
    \includegraphics[trim = 10mm 10mm 2mm 2mm, clip,width=0.12\linewidth]{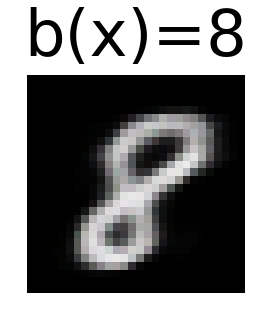}
    \caption{Exemplars \& counter-exemplars.}
    \label{fig:exemplars_mnist}
    \end{minipage}
\end{figure}

\begin{figure}[t]
    \centering
    \begin{minipage}{.48\textwidth}
    \includegraphics[trim = 10mm 10mm 2mm 2mm, clip,width=0.12\linewidth]{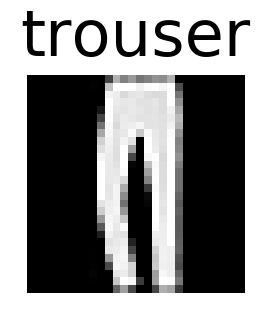}
    \includegraphics[trim = 10mm 10mm 2mm 2mm, clip,width=0.12\linewidth]{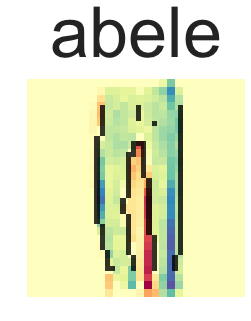}
    \includegraphics[trim = 10mm 10mm 2mm 2mm, clip,width=0.12\linewidth]{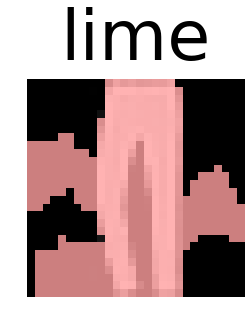}%
    \includegraphics[trim = 10mm 10mm 2mm 2mm, clip,width=0.12\linewidth]{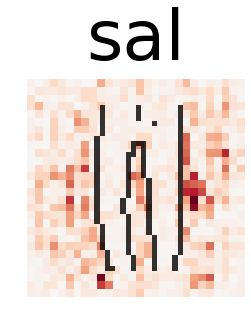}%
    \includegraphics[trim = 10mm 10mm 2mm 2mm, clip,width=0.12\linewidth]{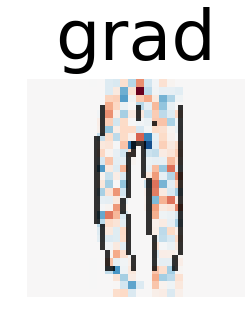}%
    \includegraphics[trim = 10mm 10mm 2mm 2mm, clip,width=0.12\linewidth]{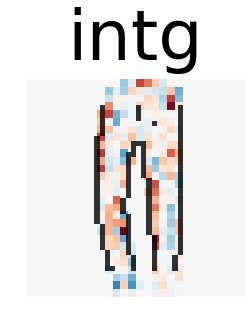}%
    \includegraphics[trim = 10mm 10mm 2mm 2mm, clip,width=0.12\linewidth]{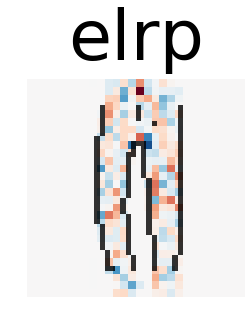}\\
    \includegraphics[trim = 10mm 10mm 2mm 2mm, clip,width=0.12\linewidth]{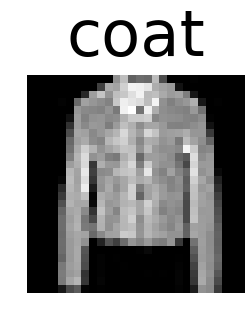}
    \includegraphics[trim = 10mm 10mm 2mm 2mm, clip,width=0.12\linewidth]{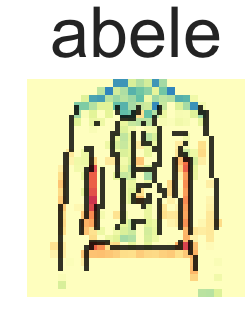}
    \includegraphics[trim = 10mm 10mm 2mm 2mm, clip,width=0.12\linewidth]{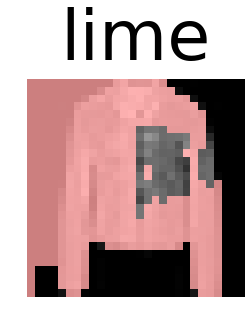}%
    \includegraphics[trim = 10mm 10mm 2mm 2mm, clip,width=0.12\linewidth]{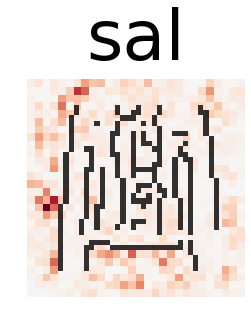}%
    \includegraphics[trim = 10mm 10mm 2mm 2mm, clip,width=0.12\linewidth]{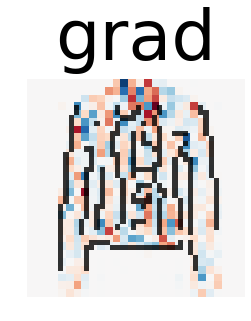}%
    \includegraphics[trim = 10mm 10mm 2mm 2mm, clip,width=0.12\linewidth]{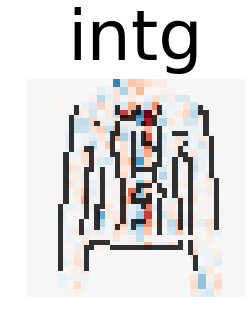}%
    \includegraphics[trim = 10mm 10mm 2mm 2mm, clip,width=0.12\linewidth]{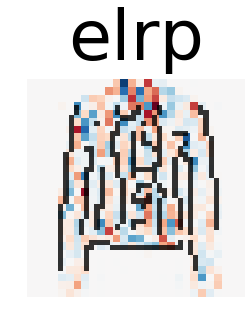}\\
    \includegraphics[trim = 10mm 10mm 2mm 2mm, clip,width=0.12\linewidth]{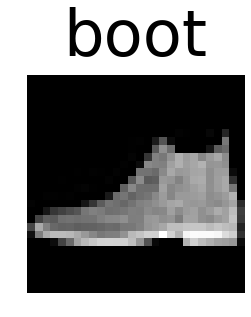}
    \includegraphics[trim = 10mm 10mm 2mm 2mm, clip,width=0.12\linewidth]{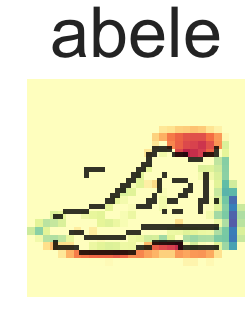}
    \includegraphics[trim = 10mm 10mm 2mm 2mm, clip,width=0.12\linewidth]{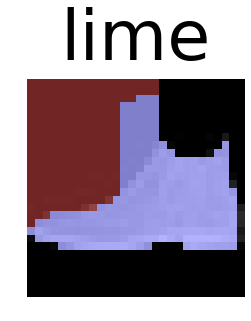}%
    \includegraphics[trim = 10mm 10mm 2mm 2mm, clip,width=0.12\linewidth]{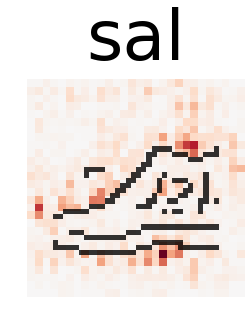}%
    \includegraphics[trim = 10mm 10mm 2mm 2mm, clip,width=0.12\linewidth]{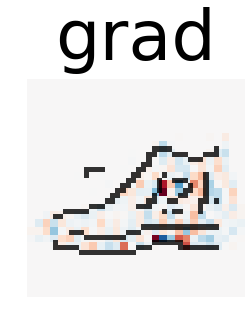}%
    \includegraphics[trim = 10mm 10mm 2mm 2mm, clip,width=0.12\linewidth]{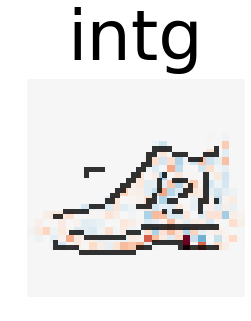}%
    \includegraphics[trim = 10mm 10mm 2mm 2mm, clip,width=0.12\linewidth]{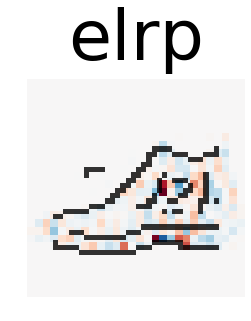}%
    \caption{Explain by saliency map \texttt{fashion}.}
    \label{fig:explanations_fashion}
    \end{minipage}%
\hspace{1mm}%
    \begin{minipage}{.48\textwidth}
    \centering
    \includegraphics[trim = 10mm 10mm 2mm 2mm, clip,width=0.12\linewidth]{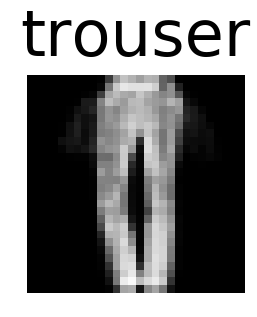}
    \includegraphics[trim = 10mm 10mm 2mm 2mm, clip,width=0.12\linewidth]{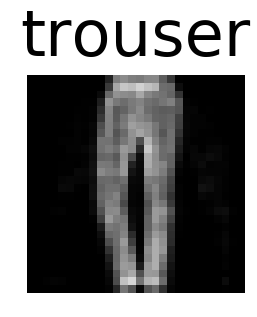}
    \includegraphics[trim = 10mm 10mm 2mm 2mm, clip,width=0.12\linewidth]{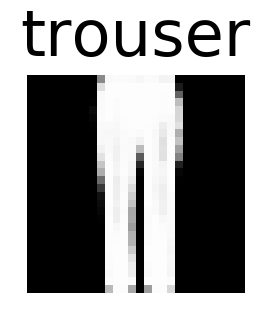}
    \hspace{1mm}
    \includegraphics[trim = 10mm 10mm 2mm 2mm, clip,width=0.12\linewidth]{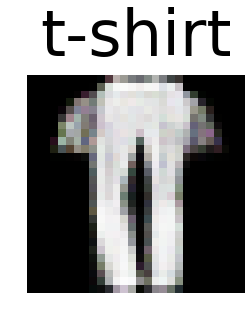}
    \includegraphics[trim = 10mm 10mm 2mm 2mm, clip,width=0.12\linewidth]{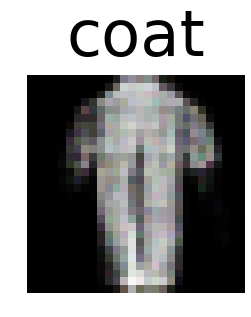}\\
    \includegraphics[trim = 10mm 10mm 2mm 2mm, clip,width=0.12\linewidth]{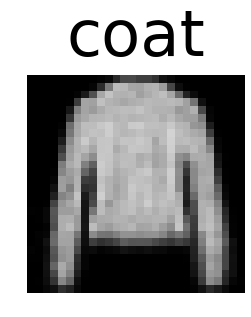}%
    \includegraphics[trim = 10mm 10mm 2mm 2mm, clip,width=0.12\linewidth]{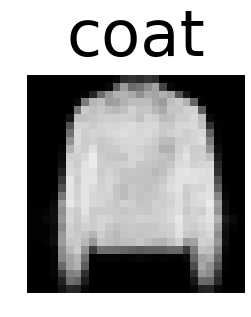}
    \includegraphics[trim = 10mm 10mm 2mm 2mm, clip,width=0.12\linewidth]{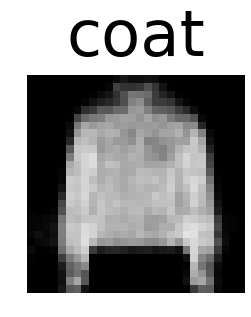}
    \hspace{1mm}
    \includegraphics[trim = 10mm 10mm 2mm 2mm, clip,width=0.12\linewidth]{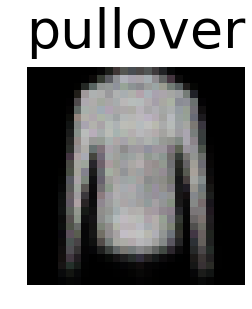}
    \includegraphics[trim = 10mm 10mm 2mm 2mm, clip,width=0.12\linewidth]{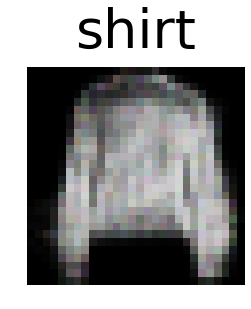}\\
    \includegraphics[trim = 10mm 10mm 2mm 2mm, clip,width=0.12\linewidth]{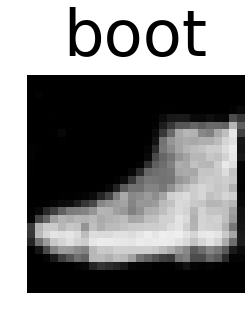}
    \includegraphics[trim = 10mm 10mm 2mm 2mm, clip,width=0.12\linewidth]{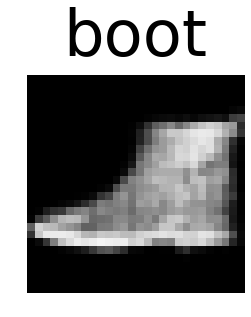}
    \includegraphics[trim = 10mm 10mm 2mm 2mm, clip,width=0.12\linewidth]{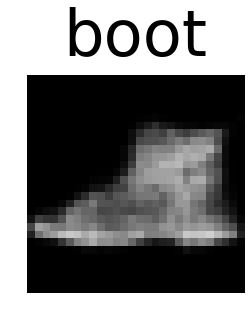}
    \hspace{1mm}%
    \includegraphics[trim = 10mm 10mm 2mm 2mm, clip,width=0.12\linewidth]{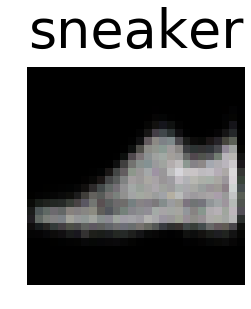}
    \includegraphics[trim = 10mm 10mm 2mm 2mm, clip,width=0.12\linewidth]{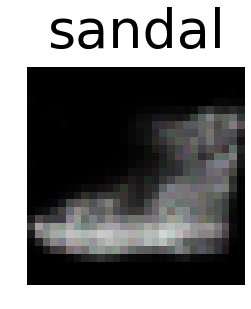}
    \caption{Exemplars \& counter-exemplars.}
    \label{fig:exemplars_fashion}
    \end{minipage}
\end{figure}

\subsection{Saliency Map, Exemplars and Counter-Exemplars}
Before assessing quantitatively the effectiveness of the compared methods, we visually analyze their outcomes.
We report explanations of the DNNs for the \texttt{mnist} and \texttt{fashion} datasets in Fig.~\ref{fig:explanations_mnist} and Fig.~\ref{fig:explanations_fashion} respectively\footnote{Best view in color. Black lines are not part of the explanation, they only highlight borders. We do not report 
explanations for \texttt{cifar10} and for RF for the sake of space.}.
The first column contains the image to explain $x$ together with the label provided by the black box $b$, while the second column contains the saliency maps provided by \abele.
Since they are derived from the difference between the image $x$ and its exemplars, we indicate with yellow color the areas that are common between $x$ and the exemplars $\widetilde{H}_e$, with red color the areas contained only in the exemplars and blue color the areas contained only in $x$.
This means that yellow areas must remain unchanged to obtain the same label $b(x)$, while red and blue areas can change without impacting the black box decision.
In particular, with respect to $x$, an image obtaining the same label can be darker in blue areas and lighter in red areas.
In other words, blue and red areas express the boundaries that can be varied, and for which the class remains unchanged.
For example, with this type of saliency map we can understand that a \textit{nine} may have a more compact circle, a \textit{zero} may be more inclined (Fig.~\ref{fig:explanations_mnist}), a \textit{coat} may have no space between the sleeves and the body, and that a \textit{boot} may have a higher neck (Fig.~\ref{fig:explanations_fashion}).
Moreover, we can notice how, besides the background, there are some ``essential'' yellow areas within the main figure that can not be different from $x$: e.g.~the leg of the \textit{nine}, the crossed lines of the \textit{four}, the space between the two \textit{trousers}.

The rest of the columns in Fig.~\ref{fig:explanations_mnist} and~\ref{fig:explanations_fashion} contain the explanations of the competitors: red areas contribute positively to the black box outcome, blue areas contribute negatively.
For \lime's explanations, nearly all the content of the image is part of the saliency areas\footnote{This effect is probably due to the figure segmentation performed by \lime.}.
In addition, the areas have either completely positive or completely negative contributions.
These aspects can be not very convincing for a \lime\ user.
On the other hand, the {\scshape dex} methods return scattered red and blue points which can also be very close to each other and are not clustered into areas.
It is not clear how a user could understand the black box outcome decision process from this kind of explanation.

\begin{figure}[t]
    \centering
    \includegraphics[trim = 10mm 10mm 2mm 2mm, clip,width=0.35\linewidth]{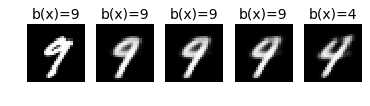}%
    \hspace{5mm}%
    \includegraphics[trim = 10mm 10mm 2mm 2mm, clip,width=0.35\linewidth]{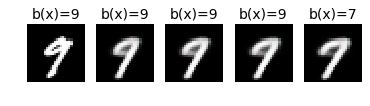}\\
    \includegraphics[trim = 10mm 10mm 2mm 2mm, clip,width=0.35\linewidth]{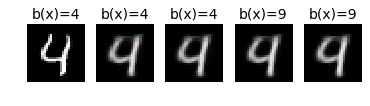}%
    \hspace{5mm}%
    \includegraphics[trim = 10mm 10mm 2mm 2mm, clip,width=0.35\linewidth]{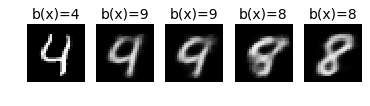}\\
    \vspace{2mm}%
    \includegraphics[trim = 10mm 10mm 2mm 8mm, clip,width=0.35\linewidth]{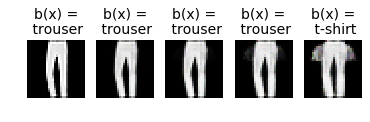}%
    \hspace{5mm}%
    \includegraphics[trim = 10mm 10mm 2mm 8mm, clip,width=0.35\linewidth]{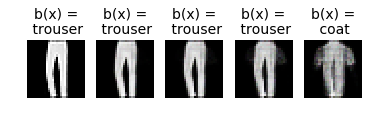}\\
    \includegraphics[trim = 10mm 10mm 2mm 8mm, clip,width=0.35\linewidth]{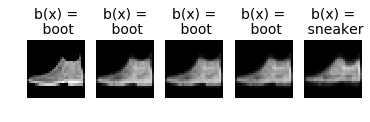}%
    \hspace{5mm}%
    \includegraphics[trim = 10mm 10mm 2mm 8mm, clip,width=0.35\linewidth]{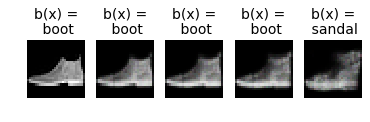}
\caption{Interpolation from the image to explain $x$ to one of its counter-exemplars $\widetilde{h_c}$. 
}
    \label{fig:counter_exemplars_interp}
\end{figure}

Since the \abele's explanations also provide exemplars and counter-exemplars, they can also be visually analyzed by a user for understanding which are possible similar instances leading to the same outcome or to a different one. 
For each instance explained in Fig.~\ref{fig:explanations_mnist} and~\ref{fig:explanations_fashion}, we show three exemplars and two counter-exemplars for the \texttt{mnist} and \texttt{fashion} datasets in Fig.~\ref{fig:exemplars_mnist} and~\ref{fig:exemplars_fashion}, respectively.
Observing these images we can notice how the label \textit{nine} is assigned to images very close to a \textit{four} (Fig.~\ref{fig:exemplars_mnist}, $1^{st}$ row, $2^{nd}$ column) but until the upper part of the circle remains connected, it is still classified as a \textit{nine}.
On the other hand, looking at counter-exemplars, if the upper part of the circle has a hole or the lower part is not thick enough, then the black box labels them as a \textit{four} and a \textit{seven}, respectively.
We highlight similar phenomena for other instances: e.g. a \textit{boot} with a neck not well defined is labeled as a \textit{sneaker} (Fig.~\ref{fig:exemplars_fashion}).

To gain further insights on the counter-exemplars, inspired by~\cite{spinner2018towards}, we exploit the latent representations to visually understand how the black box labeling changes w.r.t. real images.
In Fig.~\ref{fig:counter_exemplars_interp} we show, for some instances previously analyzed, how they can be changed to move from the original label to the counter-factual label.
We realize this change in the class through the latent representations $z$ and $h_c$ of the image to explain $x$ and of the counter-exemplar $\widetilde{h}_c$, respectively.
Given $z$ and $h_c$, we generate through linear interpolation in the latent feature space intermediate latent representations $z {<} h^{(i)}_c {<} h_c$ respecting the latent decision or counter-factual rules.
Finally, using the $\mathit{decoder}$, we obtain the intermediate images $\widetilde{h}^{(i)}_c$.
This convincing and useful explanation analysis is achieved thanks to \abele's ability to deal with both real and latent feature spaces, and to the application of latent rules to real images which are human understandable and also clear exemplar-based explanations. 

Lastly, we observe that prototype selector methods, like \mmd\ \cite{kim2016examples} and \kmedoid\, cannot be used for the same type of  analysis 
because they lack any link with either the black box or the latent space. 
In fact, they propose as prototypes (or criticism) existing images of a given dataset.
On the other hand, \abele\ generates and does not select (counter-)exemplars respecting rules.
\begin{figure}[t]
    \centering
    \includegraphics[trim = 0mm 0mm 0mm 0mm, clip,width=0.5\linewidth]{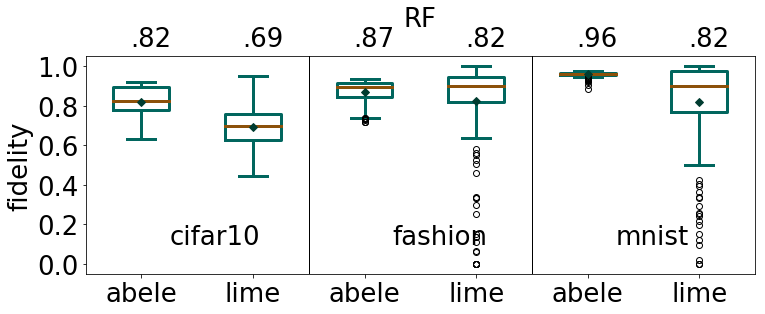}%
    \includegraphics[trim = 0mm 0mm 0mm 0mm, clip,width=0.5\linewidth]{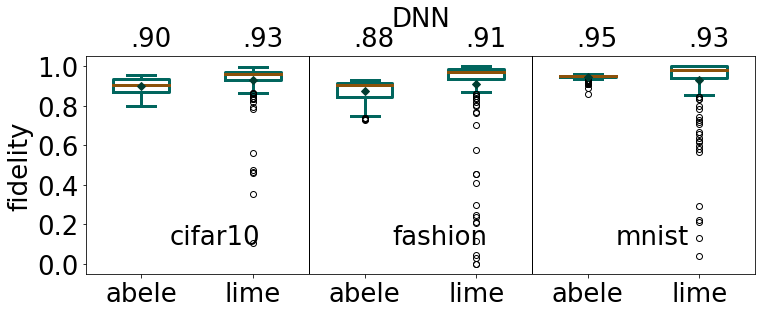}
    \caption{Box plots of \textit{fidelity}. Numbers on top: mean values (the higher the better).}
    \label{fig:fidelity}
\end{figure}

\subsection{Interpretable Classifier Fidelity}
We compare \abele\ and \lime\ in terms of 
\textit{fidelity}~\cite{guidotti2018local,doshi2017towards}, i.e.,
the ability of the local interpretable classifier $c$\footnote{A decision tree for \abele\ and a linear lasso model for \lime.} of mimicking the behavior of a black box $b$ in the local neighborhood $H$:
%
$\mathit{fidelity}(H, \widetilde{H}) = \mathit{accuracy}(b(\widetilde{H}), c(H))$.
We report the fidelity as box plots in Fig.~\ref{fig:fidelity}.
The results show that on all datasets \abele\ outperforms \lime\ with respect to the RF black box classifier.
For the DNN the interpretable classifier of \lime\ is slightly more faithful.
However, for both RF and DNN, \abele\ has a fidelity variance markedly lower than \lime, i.e., more compact box plots also without any outlier\footnote{These results confirm the experiments reported in~\cite{guidotti2018local}.}.
Since these fidelity results are statistically significant, we observe that the local interpretable classifier of \abele\ is more faithful than the one of \lime.
\begin{figure}[t]
    \centering
    \includegraphics[trim = 0mm 0mm 0mm 0mm, clip,width=0.33\linewidth]{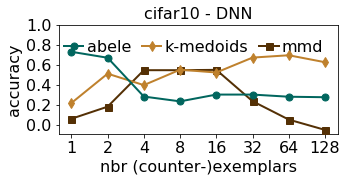}%
    \includegraphics[trim = 0mm 0mm 0mm 0mm, clip,width=0.33\linewidth]{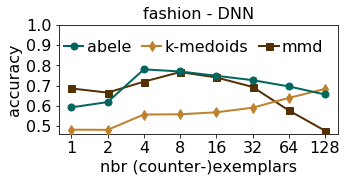}%
    \includegraphics[trim = 0mm 0mm 0mm 0mm, clip,width=0.33\linewidth]{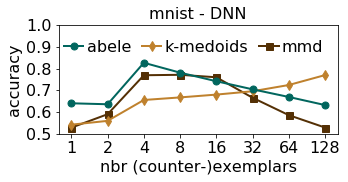} 
    \caption{1-NN exemplar classifier accuracy varying the number of (counter-)exemplars.}
   
    \label{fig:knn}
\end{figure}

\subsection{Nearest Exemplar Classifier}
The goal of \abele\ is to provide useful exemplars and counter-exemplars as explanations.
However, since we could not validate them with an experiment involving humans, inspired by~\cite{kim2016examples}, we tested their effectiveness by adopting memory-based machine learning techniques such as the k-nearest neighbor classifier~\cite{bien2011prototype} (k-NN).
This kind of experiment provides an objective and indirect evaluation of the quality of exemplars and counter-exemplars.
In the following experiment we generated $n$ exemplars and counter-exemplars with \abele, and we selected $n$ prototypes and criticisms using \mmd~\cite{kim2016examples} and
\kmedoid~\cite{bien2011prototype}.
Then, we employ a 1-NN model to classify unseen instances using these exemplars and prototypes.
The classification accuracy of the 1-NN models trained with exemplars and counter-exemplars generated to explain the DNN reported in Fig.~\ref{fig:knn} is comparable among the various methods\footnote{The \abele\ method achieves similar results for RF not reported due to lack of space.}.
In particular, we observe that when the number of exemplars is low ($1 {\leq} n {\leq} 4$), \abele\ outperforms \mmd\ and \kmedoid.
This effect reveals that, on the one hand, just a few exemplars and counter-exemplars \textit{generated} by \abele\ are good for recognizing the real label, but if the number increases the 1-NN is getting confused.
On the other hand, \mmd\ is more effective when the number of prototypes and criticisms is higher: it \textit{selects} a good set of images for the 1-NN classifier. 

\begin{figure}[t]
    \centering
    \includegraphics[trim = 0mm 0mm 0mm 0mm, clip,width=0.33\linewidth]{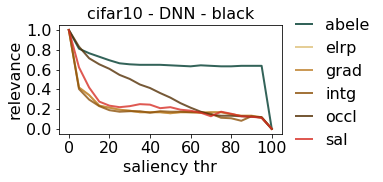}%
    \includegraphics[trim = 0mm 0mm 0mm 0mm, clip,width=0.33\linewidth]{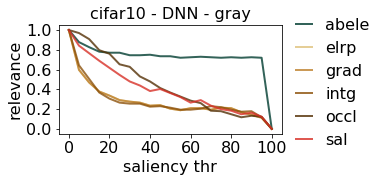}%
    \includegraphics[trim = 0mm 0mm 0mm 0mm, clip,width=0.33\linewidth]{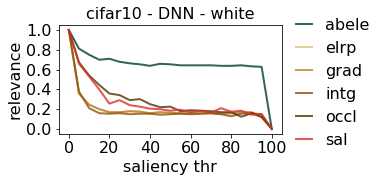}\\
    \includegraphics[trim = 0mm 0mm 0mm 0mm, clip,width=0.33\linewidth]{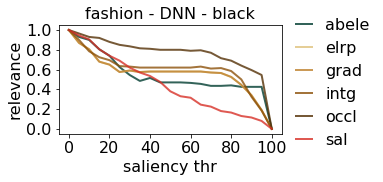}%
    \includegraphics[trim = 0mm 0mm 0mm 0mm, clip,width=0.33\linewidth]{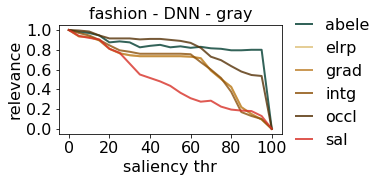}%
    \includegraphics[trim = 0mm 0mm 0mm 0mm, clip,width=0.33\linewidth]{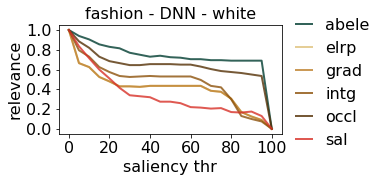}\\
    \includegraphics[trim = 0mm 0mm 0mm 0mm, clip,width=0.33\linewidth]{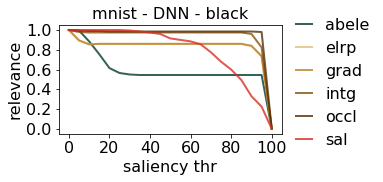}%
    \includegraphics[trim = 0mm 0mm 0mm 0mm, clip,width=0.33\linewidth]{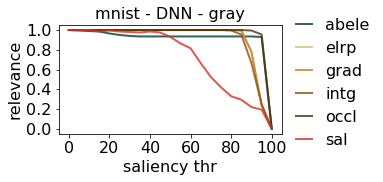}%
    \includegraphics[trim = 0mm 0mm 0mm 0mm, clip,width=0.33\linewidth]{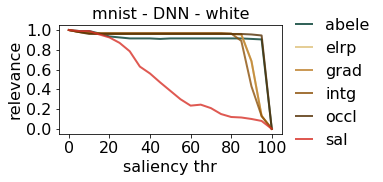}
    \caption{Relevance analysis varying the percentile threshold $\tau$ (the higher the better).}
    \label{fig:relevancy}
\end{figure}

\begin{figure}[t]
    \centering
    \begin{minipage}{.48\textwidth}
    \centering
    \includegraphics[trim = 10mm 115mm 2mm 2mm, clip,width=0.12\linewidth]{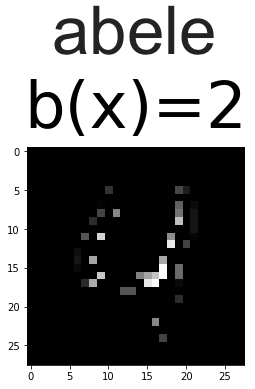}
    \includegraphics[trim = 10mm 115mm 2mm 2mm, clip,width=0.14\linewidth]{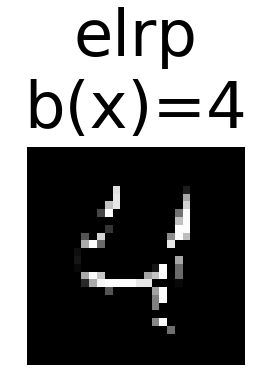}
    \includegraphics[trim = 10mm 115mm 2mm 2mm, clip,width=0.14\linewidth]{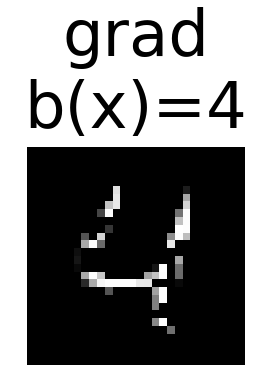}
    \includegraphics[trim = 10mm 115mm 2mm 2mm, clip,width=0.14\linewidth]{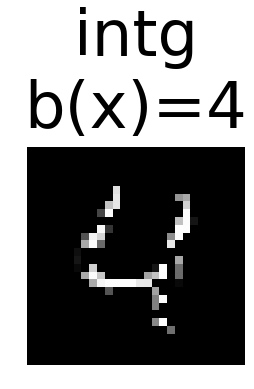}
    \includegraphics[trim = 10mm 115mm 2mm 2mm, clip,width=0.14\linewidth]{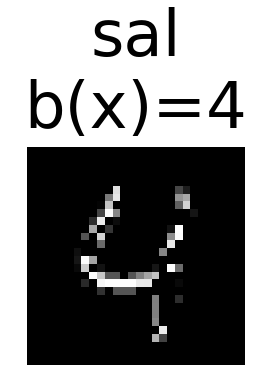}\\
    \includegraphics[trim = 10mm 10mm 2mm 25mm, clip,width=0.14\linewidth]{fig/relevancy_mnist_DNN_adele_4_black.png}
    \includegraphics[trim = 10mm 10mm 2mm 25mm, clip,width=0.14\linewidth]{fig/relevancy_mnist_DNN_elrp_4_black.png}
    \includegraphics[trim = 10mm 10mm 2mm 25mm, clip,width=0.14\linewidth]{fig/relevancy_mnist_DNN_gradinput_4_black.png}
    \includegraphics[trim = 10mm 10mm 2mm 25mm, clip,width=0.14\linewidth]{fig/relevancy_mnist_DNN_intgrad_4_black.png}
    \includegraphics[trim = 10mm 10mm 2mm 25mm, clip,width=0.14\linewidth]{fig/relevancy_mnist_DNN_saliency_4_black.png}\\
    \includegraphics[trim = 10mm 10mm 2mm 25mm, clip,width=0.14\linewidth]{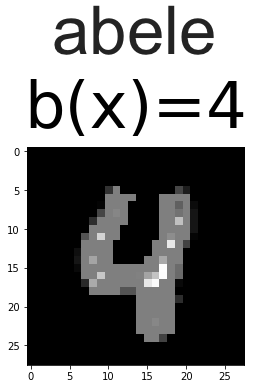}
    \includegraphics[trim = 10mm 10mm 2mm 25mm, clip,width=0.14\linewidth]{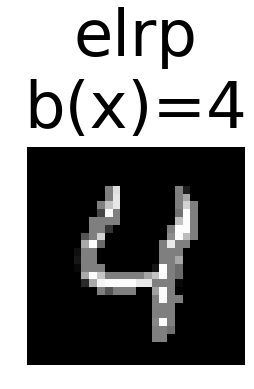}
    \includegraphics[trim = 10mm 10mm 2mm 25mm, clip,width=0.14\linewidth]{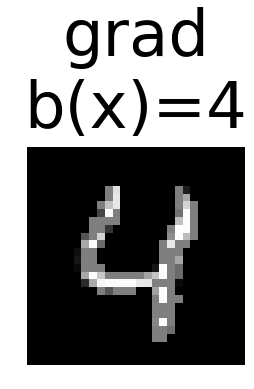}
    \includegraphics[trim = 10mm 10mm 2mm 25mm, clip,width=0.14\linewidth]{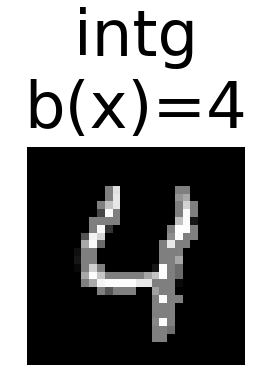}
    \includegraphics[trim = 10mm 10mm 2mm 25mm, clip,width=0.14\linewidth]{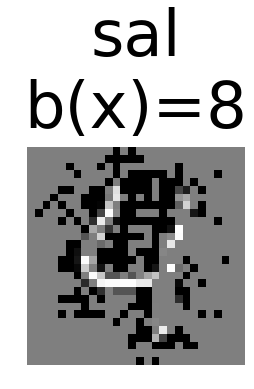}\\
    \includegraphics[trim = 10mm 10mm 2mm 25mm, clip,width=0.14\linewidth]{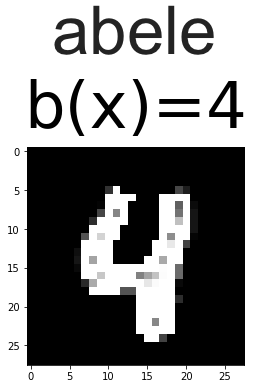}
    \includegraphics[trim = 10mm 10mm 2mm 25mm, clip,width=0.14\linewidth]{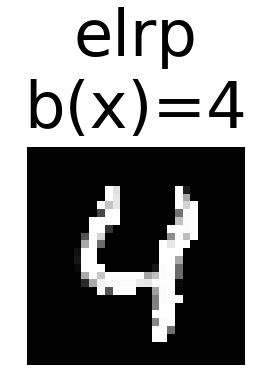}
    \includegraphics[trim = 10mm 10mm 2mm 25mm, clip,width=0.14\linewidth]{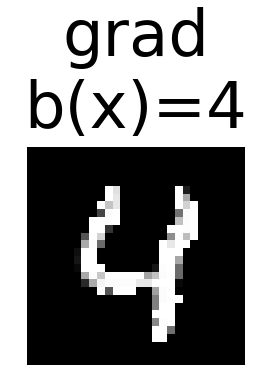}
    \includegraphics[trim = 10mm 10mm 2mm 25mm, clip,width=0.14\linewidth]{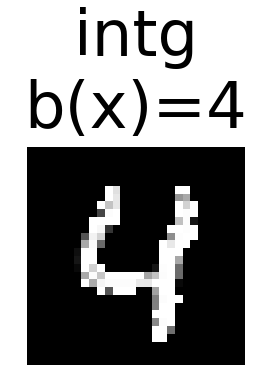}
    \includegraphics[trim = 10mm 10mm 2mm 25mm, clip,width=0.14\linewidth]{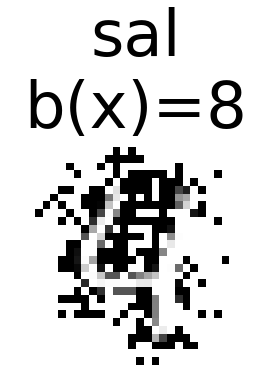}
    \end{minipage}
    \begin{minipage}{.48\textwidth}
    \centering
    \includegraphics[trim = 10mm 115mm 2mm 2mm, clip,width=0.14\linewidth]{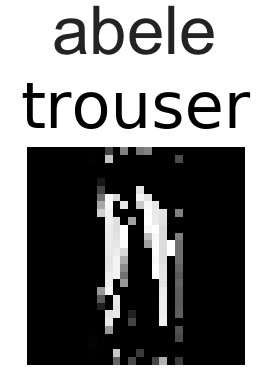}
    \includegraphics[trim = 10mm 115mm 2mm 2mm, clip,width=0.14\linewidth]{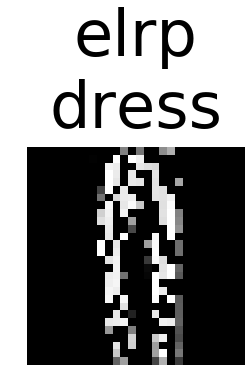}
    \includegraphics[trim = 10mm 115mm 2mm 2mm, clip,width=0.14\linewidth]{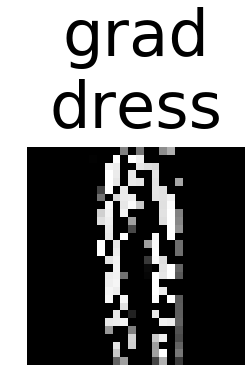}
    \includegraphics[trim = 10mm 115mm 2mm 2mm, clip,width=0.14\linewidth]{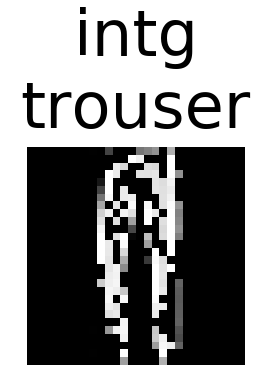}
    \includegraphics[trim = 10mm 115mm 2mm 2mm, clip,width=0.14\linewidth]{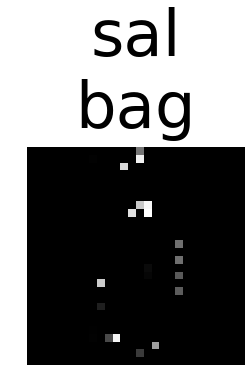}\\
    \includegraphics[trim = 10mm 10mm 2mm 25mm, clip,width=0.14\linewidth]{fig/relevancy_fashion_DNN_adele_1_black.png}
    \includegraphics[trim = 10mm 10mm 2mm 25mm, clip,width=0.14\linewidth]{fig/relevancy_fashion_DNN_elrp_1_black.png}
    \includegraphics[trim = 10mm 10mm 2mm 25mm, clip,width=0.14\linewidth]{fig/relevancy_fashion_DNN_gradinput_1_black.png}
    \includegraphics[trim = 10mm 10mm 2mm 25mm, clip,width=0.14\linewidth]{fig/relevancy_fashion_DNN_intgrad_1_black.png}
    \includegraphics[trim = 10mm 10mm 2mm 25mm, clip,width=0.14\linewidth]{fig/relevancy_fashion_DNN_saliency_1_black.png}\\
    \includegraphics[trim = 10mm 10mm 2mm 25mm, clip,width=0.14\linewidth]{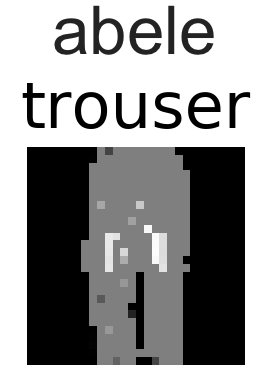}
    \includegraphics[trim = 10mm 10mm 2mm 25mm, clip,width=0.14\linewidth]{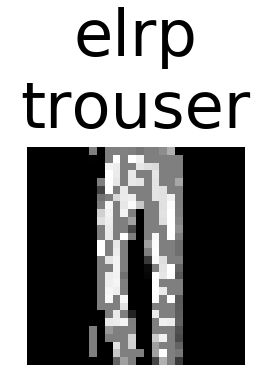}
    \includegraphics[trim = 10mm 10mm 2mm 25mm, clip,width=0.14\linewidth]{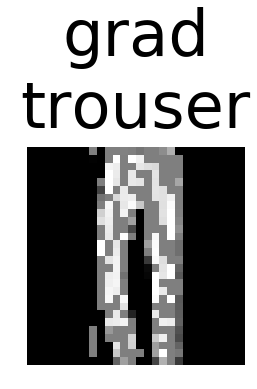}
    \includegraphics[trim = 10mm 10mm 2mm 25mm, clip,width=0.14\linewidth]{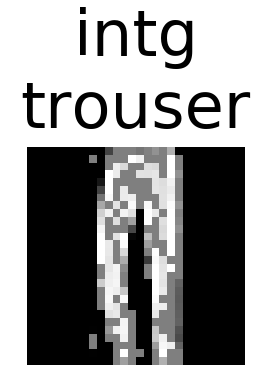}
    \includegraphics[trim = 10mm 10mm 2mm 25mm, clip,width=0.14\linewidth]{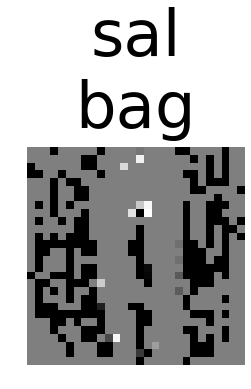}\\
    \includegraphics[trim = 10mm 10mm 2mm 25mm, clip,width=0.14\linewidth]{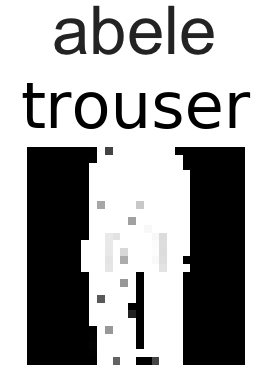}
    \includegraphics[trim = 10mm 10mm 2mm 25mm, clip,width=0.14\linewidth]{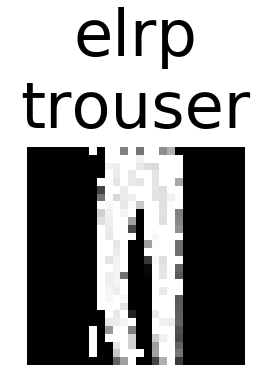}
    \includegraphics[trim = 10mm 10mm 2mm 25mm, clip,width=0.14\linewidth]{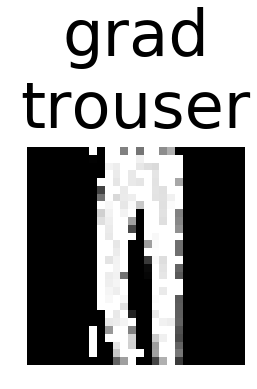}
    \includegraphics[trim = 10mm 10mm 2mm 25mm, clip,width=0.14\linewidth]{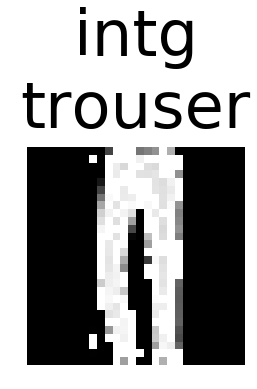}
    \includegraphics[trim = 10mm 10mm 2mm 25mm, clip,width=0.14\linewidth]{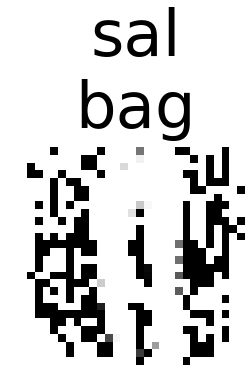}
    \end{minipage}
    \caption{Images masked with \textit{black}, \textit{gray} and \textit{white} having pixels with saliency for DNN lower than $\tau=70\%$ for the explanations of \textit{four} and \textit{trouser} in Fig.~\ref{fig:explanations_mnist} and~\ref{fig:explanations_fashion}.}
    \label{fig:relevancy_examples}
\end{figure}

\subsection{Relevance Evaluation}
We evaluate the effectiveness of \abele\ by partly masking the image to explain $x$.
According to~\cite{hara2018maximally}, although a part of $x$ is masked, $b(x)$ should remain unchanged as long as relevant parts of $x$ remain unmasked.
To quantitatively measure this aspect, we define the $\mathit{relevance}$ metric as the ratio of images in $X$ for which the masking of relevant parts does not impact on the black box decision.
Let ${E{=}\{e_1, \mydots, e_n\}}$ be the set of explanations for the instances ${X{=}\{x_1, \mydots, x_n\}}$. 
We identify with $x^{\{e, \tau\}}_m$ the masked version of $x$ with respect to the explanation $e$ and a threshold mask $\tau$.
Then, the explanation \textit{relevance} is defined as:
%
$\mathit{relevance}_{\tau}(X, E) = \vert \{x \:|\: b(x) = b(x^{\{e, \tau\}}_m) \; \forall \langle x, e \rangle \in \langle X, E \rangle \} \vert \; / \; \vert X \vert$.
%
The masking $x^{\{e, \tau\}}_m$ is got by changing the pixels of $x$ having a value in the saliency map $s \in e$ smaller than the $\tau$ percentile of the set of values in the saliency map itself. 
These pixels 
are substituted with the color \textit{0}, \textit{127} or \textit{255}, i.e.~\textit{black}, \textit{gray} or \textit{white}.
A low number of black box outcome changes means that the explainer successfully identifies \textit{relevant} parts of the images, i.e., parts having a high relevance. Fig.~\ref{fig:relevancy} shows the \textit{relevance} for the DNN\footnote{The \abele\ method achieves similar results for RF not reported due to lack of space.} varying the percentile of the threshold from 0 to 100. 
The \abele\ method is the most resistant to image masking in \texttt{cifar10} regardless of the color used.
For the other datasets we observe a different behavior depending on the masking color used: \abele\ is among the best performer if the masking color is \textit{white} or \textit{gray}, while when the mask color is \textit{black}, \abele's relevance is in line with those of the competitors for \texttt{fashion} and it is not good for \texttt{mnist}.
This effect depends on the masking color but also on the different definitions of saliency map. 
Indeed, as previously discussed, depending on the explainer, a saliency map can provide different knowledge.
However, we can state that \abele\ successfully identifies relevant parts of the image contributing to the classification.

For each method and for each masking color, Fig.~\ref{fig:relevancy_examples} shows the effect of the masking on a sample from \texttt{mnist} and another from \texttt{fashion}.
It is interesting to notice how for the \sal\ approach a large part of the image is quite relevant, causing a different black box outcome (reported on the top of each image).
As already observed previously, a peculiarity of \abele\ is that the saliency areas are more connected and larger than those of the other methods.
Therefore, given a percentile threshold $\tau$, the masking operation tends to mask more contiguous and bigger areas of the image while maintaining the same black box labeling. 

\begin{table}[t]
\centering
\setlength{\tabcolsep}{0.7mm}
\begin{tabular}{|c|ccccccc|}
\hline
dataset & \abele & \elrp & \grad & \intg & \lime & \occ & \sal \\
\hline
 \tt  cifar10 & $.575 \pm .10$ & $.542 \pm .08$ & $.542 \pm .08$ & $.532 \pm .11$ & $1.919 \pm .25$ & $1.08 \pm .23$ & $.471 \pm .05$ \\
 \tt fashion & $.451 \pm .06$ & $.492 \pm .10$ & $.492 \pm .10$ & $.561 \pm .17$ & $1.618 \pm .16$ & $.904 \pm .23$ & $.413 \pm .03$ \\
 \tt mnist & $.380 \pm .03$ & $.740 \pm .21$ & $.740 \pm .21$ & $.789 \pm .22$ & $1.475 \pm .14$ & $.734 \pm .21$ & $.391 \pm .03$ \\
\hline
\end{tabular}
\caption{Coherence analysis for DNN 
classifier (the lower the better).}
\label{tab:coherence}
\end{table}

\begin{table}[t]
\centering
\setlength{\tabcolsep}{0.7mm}
\begin{tabular}{|c|ccccccc|}
\hline
dataset & \abele & \elrp & \grad & \intg & \lime & \occ & \sal \\
\hline
\tt  cifar10 & $.575 \pm .10$ & $.518 \pm .08$ & $.518 \pm .08$ & $.561 \pm .10$ & $1.898 \pm .29$ & $.957 \pm .14$ & $.468 \pm .05$ \\
\tt fashion & $.455 \pm .06$ & $.490 \pm .09$ & $.490 \pm .09$ & $.554 \pm .18$ & $1.616 \pm .17$ & $.908 \pm .23$ & $.415 \pm .03$ \\
\tt mnist & $.380 \pm .04$ & $.729 \pm .21$ & $.729 \pm .21$ & $.776 \pm .22$ & $1.485 \pm .14$ & $.726 \pm .21$ & $.393 \pm .03$ \\
\hline
\end{tabular}
\caption{Stability analysis for DNN 
classifier (the lower the better).}
\label{tab:stability}
\end{table}

\begin{table}[!t]
\begin{minipage}{.5\linewidth}
\centering
\setlength{\tabcolsep}{1.2mm}
\begin{tabular}{|c|cc|}
\hline
dataset & \abele & \lime \\
\hline
\tt cifar10 & $.794 \pm .34$ & $1.692 \pm .32$ \\
\tt fashion & $.821 \pm .37$ & $2.534 \pm .70$ \\
\tt mnist & $.568 \pm .29$ & $2.593 \pm 1.25$ \\
\hline
\end{tabular}
\end{minipage}%
\begin{minipage}{.5\linewidth}
\centering
\setlength{\tabcolsep}{1.2mm}
\begin{tabular}{|c|cc|}
\hline
dataset & \abele & \lime \\
\hline
\tt cifar10 & $.520 \pm .14$ & $1.460 \pm .23$ \\
\tt fashion & $.453 \pm .06$ & $1.464 \pm .18$ \\
\tt mnist & $.371 \pm .04$ & $1.451 \pm .17$ \\
\hline
\end{tabular}
\end{minipage}
\caption{Coherence (left) and stability (right) for RF 
classifier (the lower the better).}
\label{tab:robustness_rf}
\end{table}

\subsection{Robustness Assessment}
For gaining the trust of the user, it is crucial to analyze the stability of interpretable classifiers and explainers~\cite{guidotti2019stability} since the stability of explanations is an important requirement for interpretability~\cite{melis2018towards}.
Let ${E{=}\{e_1, \mydots, e_n\}}$ be the set of explanations for ${X{=}\{x_1, \mydots, x_n\}}$, and $\{s_1, \mydots, s_n\}$ the corresponding saliency maps.
We asses the \textit{robustness} through the local Lipschitz estimation~\cite{melis2018towards}:
$\mathit{robustness}(x) = \mathit{argmax}_{x_i \in \mathcal{N}(x)}(\lVert s_i - s \rVert_2 / \lVert x_i - x \rVert_2)$ with
$\mathcal{N}(x) = \{ x_j {\in} X \; | \; \lVert x_j - x \rVert_2 \leq \epsilon\}$.
%
Here $x$ is the image to explain and $s$ is the saliency map of its explanation $e$.
We name \textit{coherence} the explainer's ability to return similar explanations to instances labeled with the same black box outcome, i.e., similar instances. 
We name \textit{stability}, often called also \textit{sensitivity}, the capacity of an explainer of not varying an explanation in the presence of noise with respect to the explained instance.
Therefore, for coherence the set $X$ in the \textit{robustness} formula is formed by real instances, while for stability $X$ is formed by the instances to explain modified with random noise\footnote{As in~\cite{melis2018towards}, in our experiments, we use $\epsilon{=}0.1$ for $\mathcal{N}$ and we add salt and pepper noise.}. 

\begin{figure}[t]
    \centering
    \begin{minipage}{.48\textwidth}
    \includegraphics[trim = 10mm 10mm 2mm 2mm, clip,width=0.13\linewidth]{fig/exemplar_mnist_DNN_original_hrgp_9.png}
    \includegraphics[trim = 10mm 10mm 2mm 2mm, clip,width=0.13\linewidth]{fig/exemplar_mnist_DNN_adele_hrgp_9.png}
    \includegraphics[trim = 10mm 10mm 2mm 2mm, clip,width=0.13\linewidth]{fig/exemplar_mnist_DNN_lime_9.png}
    \includegraphics[trim = 10mm 10mm 2mm 2mm, clip,width=0.13\linewidth]{fig/exemplar_mnist_DNN_saliency_9.png}
    \includegraphics[trim = 10mm 10mm 2mm 2mm, clip,width=0.13\linewidth]{fig/exemplar_mnist_DNN_gradinput_9.png}
    \includegraphics[trim = 10mm 10mm 2mm 2mm, clip,width=0.13\linewidth]{fig/exemplar_mnist_DNN_intgrad_9.png}
    \includegraphics[trim = 10mm 10mm 2mm 2mm, clip,width=0.13\linewidth]{fig/exemplar_mnist_DNN_elrp_9.png}\\
    \includegraphics[trim = 10mm 10mm 2mm 2mm, clip,width=0.13\linewidth]{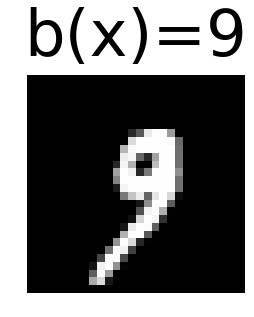}
    \includegraphics[trim = 10mm 10mm 2mm 2mm, clip,width=0.13\linewidth]{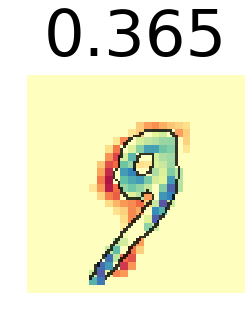}
    \includegraphics[trim = 10mm 10mm 2mm 2mm, clip,width=0.13\linewidth]{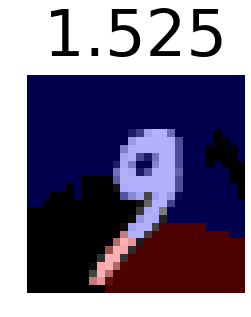}
    \includegraphics[trim = 10mm 10mm 2mm 2mm, clip,width=0.13\linewidth]{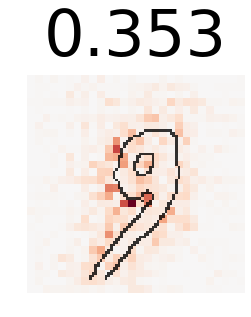}
    \includegraphics[trim = 10mm 10mm 2mm 2mm, clip,width=0.13\linewidth]{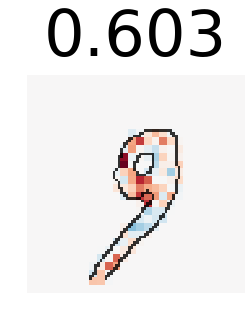}
    \includegraphics[trim = 10mm 10mm 2mm 2mm, clip,width=0.13\linewidth]{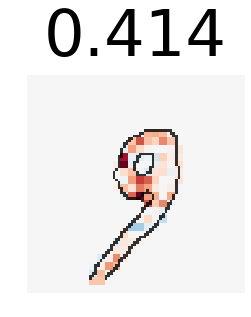}
    \includegraphics[trim = 10mm 10mm 2mm 2mm, clip,width=0.13\linewidth]{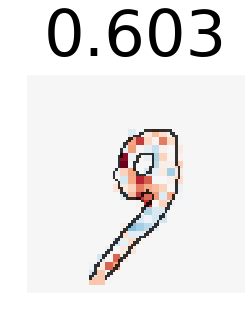}
    \end{minipage}
    \hspace{2mm}
    \begin{minipage}{.48\textwidth}
    \includegraphics[trim = 10mm 10mm 2mm 2mm, clip,width=0.13\linewidth]{fig/exemplar_fashion_DNN_original_hrgp_9.png}
    \includegraphics[trim = 10mm 10mm 2mm 2mm, clip,width=0.13\linewidth]{fig/exemplar_fashion_DNN_adele_hrgp_9.png}
    \includegraphics[trim = 10mm 10mm 2mm 2mm, clip,width=0.13\linewidth]{fig/exemplar_fashion_DNN_lime_9.png}
    \includegraphics[trim = 10mm 10mm 2mm 2mm, clip,width=0.13\linewidth]{fig/exemplar_fashion_DNN_saliency_9.png}
    \includegraphics[trim = 10mm 10mm 2mm 2mm, clip,width=0.13\linewidth]{fig/exemplar_fashion_DNN_gradinput_9.png}
    \includegraphics[trim = 10mm 10mm 2mm 2mm, clip,width=0.13\linewidth]{fig/exemplar_fashion_DNN_intgrad_9.png}
    \includegraphics[trim = 10mm 10mm 2mm 2mm, clip,width=0.13\linewidth]{fig/exemplar_fashion_DNN_elrp_9.png}\\
    \includegraphics[trim = 10mm 10mm 2mm 2mm, clip,width=0.13\linewidth]{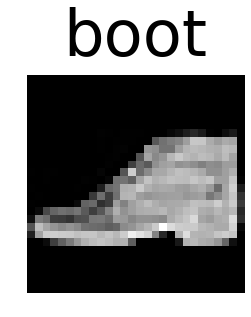}
    \includegraphics[trim = 10mm 10mm 2mm 2mm, clip,width=0.13\linewidth]{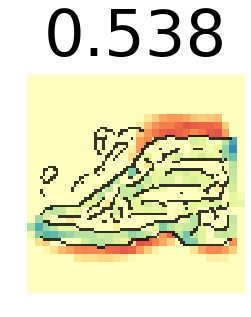}
    \includegraphics[trim = 10mm 10mm 2mm 2mm, clip,width=0.13\linewidth]{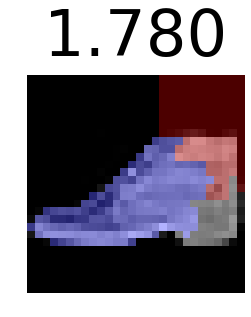}
    \includegraphics[trim = 10mm 10mm 2mm 2mm, clip,width=0.13\linewidth]{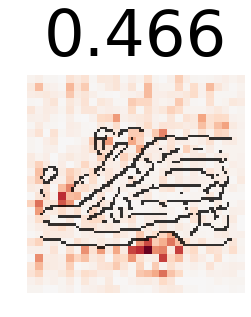}
    \includegraphics[trim = 10mm 10mm 2mm 2mm, clip,width=0.13\linewidth]{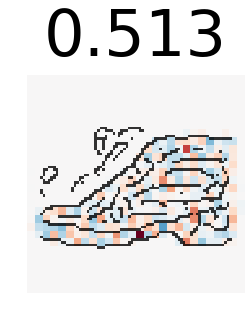}
    \includegraphics[trim = 10mm 10mm 2mm 2mm, clip,width=0.13\linewidth]{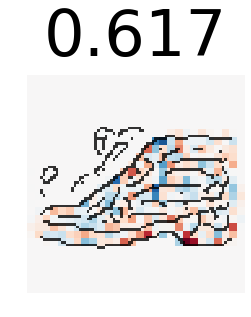}
    \includegraphics[trim = 10mm 10mm 2mm 2mm, clip,width=0.13\linewidth]{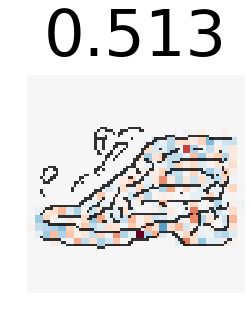}
    \end{minipage}
    \caption{Saliency maps for \texttt{mnist} (left) and \texttt{fashion} (right) comparing two images with the same DNN outcome; numbers on the top are the \textit{coherence} (the lower the better).}
    \label{fig:coherence}
\end{figure}

\begin{figure}[t]
    \centering
    \begin{minipage}{.48\textwidth}
    \includegraphics[trim = 10mm 10mm 2mm 2mm, clip,width=0.13\linewidth]{fig/exemplar_mnist_DNN_original_hrgp_0.png}
    \includegraphics[trim = 10mm 10mm 2mm 2mm, clip,width=0.13\linewidth]{fig/exemplar_mnist_DNN_adele_hrgp_0.png}
    \includegraphics[trim = 10mm 10mm 2mm 2mm, clip,width=0.13\linewidth]{fig/exemplar_mnist_DNN_lime_0.png}
    \includegraphics[trim = 10mm 10mm 2mm 2mm, clip,width=0.13\linewidth]{fig/exemplar_mnist_DNN_saliency_0.png}
    \includegraphics[trim = 10mm 10mm 2mm 2mm, clip,width=0.13\linewidth]{fig/exemplar_mnist_DNN_gradinput_0.png}
    \includegraphics[trim = 10mm 10mm 2mm 2mm, clip,width=0.13\linewidth]{fig/exemplar_mnist_DNN_intgrad_0.png}
    \includegraphics[trim = 10mm 10mm 2mm 2mm, clip,width=0.13\linewidth]{fig/exemplar_mnist_DNN_elrp_0.png}\\
    \includegraphics[trim = 10mm 10mm 2mm 2mm, clip,width=0.13\linewidth]{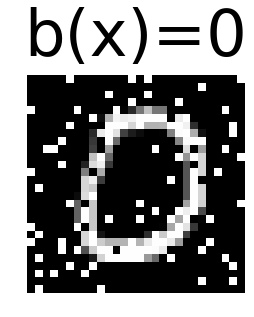}
    \includegraphics[trim = 10mm 10mm 2mm 2mm, clip,width=0.13\linewidth]{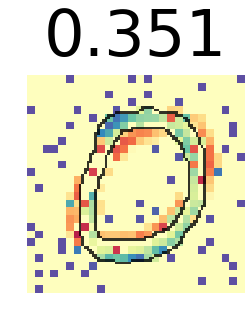}
    \includegraphics[trim = 10mm 10mm 2mm 2mm, clip,width=0.13\linewidth]{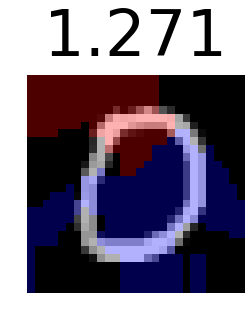}
    \includegraphics[trim = 10mm 10mm 2mm 2mm, clip,width=0.13\linewidth]{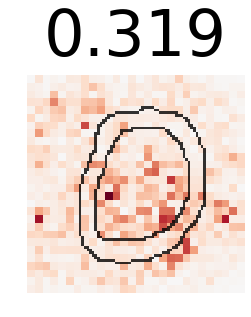}
    \includegraphics[trim = 10mm 10mm 2mm 2mm, clip,width=0.13\linewidth]{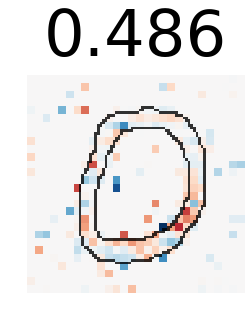}
    \includegraphics[trim = 10mm 10mm 2mm 2mm, clip,width=0.13\linewidth]{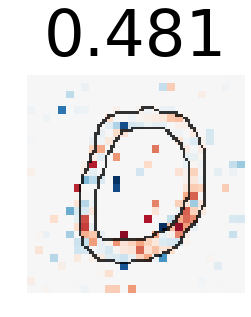}
    \includegraphics[trim = 10mm 10mm 2mm 2mm, clip,width=0.13\linewidth]{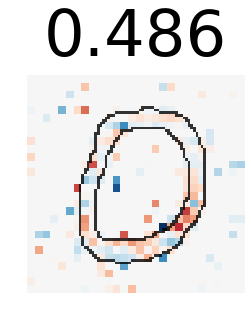}
    \end{minipage}
    \hspace{2mm}
    \begin{minipage}{.48\textwidth}
    \includegraphics[trim = 10mm 10mm 2mm 2mm, clip,width=0.13\linewidth]{fig/exemplar_fashion_DNN_original_hrgp_4.png}
    \includegraphics[trim = 10mm 10mm 2mm 2mm, clip,width=0.13\linewidth]{fig/exemplar_fashion_DNN_adele_hrgp_4.png}
    \includegraphics[trim = 10mm 10mm 2mm 2mm, clip,width=0.13\linewidth]{fig/exemplar_fashion_DNN_lime_4.png}
    \includegraphics[trim = 10mm 10mm 2mm 2mm, clip,width=0.13\linewidth]{fig/exemplar_fashion_DNN_saliency_4.png}
    \includegraphics[trim = 10mm 10mm 2mm 2mm, clip,width=0.13\linewidth]{fig/exemplar_fashion_DNN_gradinput_4.png}
    \includegraphics[trim = 10mm 10mm 2mm 2mm, clip,width=0.13\linewidth]{fig/exemplar_fashion_DNN_intgrad_4.png}
    \includegraphics[trim = 10mm 10mm 2mm 2mm, clip,width=0.13\linewidth]{fig/exemplar_fashion_DNN_elrp_4.png}\\
    \includegraphics[trim = 10mm 10mm 2mm 2mm, clip,width=0.13\linewidth]{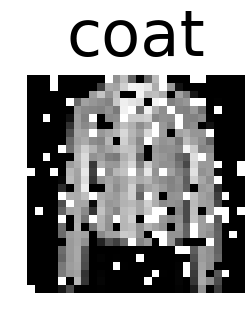}
    \includegraphics[trim = 10mm 10mm 2mm 2mm, clip,width=0.13\linewidth]{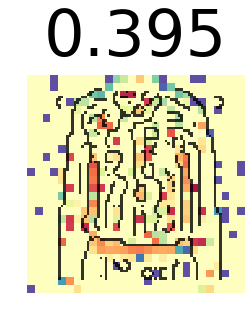}
    \includegraphics[trim = 10mm 10mm 2mm 2mm, clip,width=0.13\linewidth]{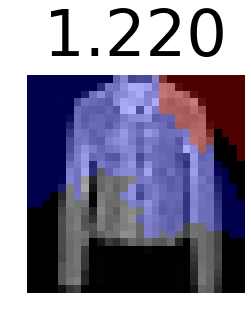}
    \includegraphics[trim = 10mm 10mm 2mm 2mm, clip,width=0.13\linewidth]{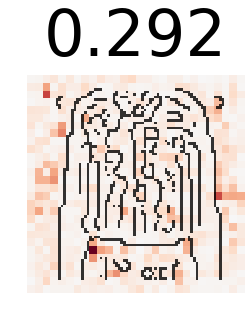}
    \includegraphics[trim = 10mm 10mm 2mm 2mm, clip,width=0.13\linewidth]{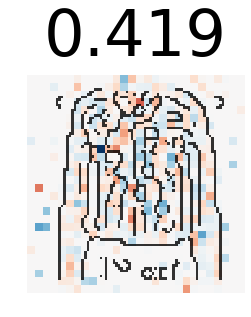}
    \includegraphics[trim = 10mm 10mm 2mm 2mm, clip,width=0.13\linewidth]{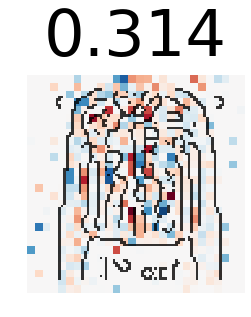}
    \includegraphics[trim = 10mm 10mm 2mm 2mm, clip,width=0.13\linewidth]{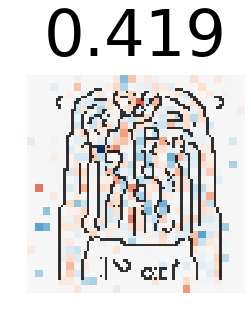}
    \end{minipage}
    \caption{Saliency maps for \texttt{mnist} (left) and \texttt{fashion} (right) comparing the original image in the first row and the modified version with salt and pepper noise but with the same DNN outcome; numbers on the top are the \textit{stability} (the lower the better).}
    \label{fig:stability}
\end{figure}

Tables~\ref{tab:coherence} and~\ref{tab:stability} report mean and standard deviation of the local Lipschitz estimations of the explainers' \textit{robustness} in terms of \textit{coherence} and \textit{stability}, respectively.
As showed in~\cite{melis2018towards}, our results confirm that \lime\ does not provide robust explanations, \grad\ and \intg\ are the best performers, and \abele\ performance is comparable to them in terms of both \textit{coherence} and \textit{stability}.
This high resilience of \abele\ is due to the usage of AAE, which is also adopted for image denoising~\cite{xie2012image}.
Table~\ref{tab:robustness_rf} shows the robustness in terms of coherence and stability for the model agnostic explainers \abele\ and \lime\ with respect to the RF.
Again, \abele\ presents a more robust behavior than \lime.
Fig.~\ref{fig:coherence} and~\ref{fig:stability} compare the saliency maps of a selected image from \texttt{mnist} and \texttt{fashion} labeled with DNN.
Numbers on the top represent the ratio in the robustness formula.
Although there is no change in the black box outcome, we can see how for some of the other explainers like \lime, \elrp, and \grad, the saliency maps vary considerably.
On the other hand, \abele's explanations remain coherent and stable.
We observe how in both \textit{nines} and \textit{boots} the yellow fundamental area does not change especially within the image's edges. 
Also the red and blue parts, that can be varied without impacting on the classification, are almost identical, e.g.~the \textit{boots}' neck and the sole in Fig.~\ref{fig:coherence}, or the top left of the \textit{zero} in Fig.~\ref{fig:stability}.

\section{Conclusion}
\label{sec:conclusion}
We have presented \abele, a local model-agnostic explainer using the latent feature space learned through an adversarial autoencoder for the neighborhood generation process.
The explanation returned by \abele\ consists of 
exemplar and counter-exemplar images, labeled 
with the class identical to, and different from, the class of the image to explain,  
and by a a saliency map, highlighting the importance of the areas of the image contributing to its classification. 
An extensive experimental comparison with state of the art methods shows that \abele\ addresses their deficiencies, and outperforms them by returning coherent, stable and faithful explanations.

The method has some limitations: it is constrained to image data and does not enable casual or logical reasoning.
Several extensions and future work are possible. 
First, we would like to investigate the effect on the explanations of changing some aspect of the AAE:
\textit{(i)} the latent dimensions $k$, \textit{(ii)} the rigidity of the $\mathit{discriminator}$ in admitting latent instances,
\textit{(iii)} the type of autoencoders (e.g.~variational autoencoders~\cite{siddharth2016inducing}).
Second, we would like to extend \abele\ to make it work on tabular data and on text. 
Third, we would employ \abele\ in a case study generating exemplars and counter-exemplars for explaining medical imaging tasks, e.g. radiography and fMRI images.
Lastly, we would conduct extrinsic interpretability evaluation of \abele. 
Human decision-making in a specific task (e.g. multiple-choice question answering) would be driven by \abele\ explanations, and these decisions 
could be objectively and quantitatively evaluated. 

\medskip \noindent
{\footnotesize\textbf{Acknowledgements.} 
This work is partially supported by the EC H2020 programme under the funding schemes: Research Infrastructures G.A. 6540\-24 \emph{SoBigData},  
G.A. 78835 \emph{Pro-Res},  
G.A. 825619 \emph{AI4EU} and 
G.A. 780754 \emph{Track\&Know}. 
The third author acknowledges the support of the Natural Sciences and Engineering Research Council of Canada and of the Ocean Frontiers Institute. 
}



%
%

%
%
%
\bibliographystyle{abbrv}
\bibliography{biblio}

\begin{thebibliography}{10}

\bibitem{bach2015pixel}
S.~Bach, A.~Binder, et~al.
\newblock On pixel-wise explanations for non-linear classifier decisions by
  layer-wise relevance propagation.
\newblock {\em PloS one}, 10(7):e0130140, 2015.

\bibitem{bien2011prototype}
J.~Bien et~al.
\newblock Prototype selection for interpretable classification.
\newblock {\em AOAS}, 2011.

\bibitem{breiman2001random}
L.~Breiman.
\newblock Random forests.
\newblock {\em Machine learning}, 45(1):5--32, 2001.

\bibitem{chen2018looks}
C.~Chen, O.~Li, A.~Barnett, J.~Su, and C.~Rudin.
\newblock This looks like that: deep learning for interpretable image
  recognition.
\newblock {\em arXiv:1806.10574}, 2018.

\bibitem{doshi2017towards}
F.~Doshi-Velez and B.~Kim.
\newblock Towards a rigorous science of interpretable machine learning.
\newblock {\em arXiv:1702.08608}, 2017.

\bibitem{escalante2018explainable}
H.~J. Escalante, S.~Escalera, I.~Guyon, et~al.
\newblock {\em Explainable and interpretable models in computer vision and
  machine learning}.
\newblock Springer, 2018.

\bibitem{fong2017interpretable}
R.~C. Fong and A.~Vedaldi.
\newblock Interpretable explanations of black boxes by meaningful perturbation.
\newblock In {\em ICCV}, pages 3429--3437, 2017.

\bibitem{frixione2012prototypes}
M.~Frixione et~al.
\newblock Prototypes vs exemplars in concept representation. {KEOD}, 2012.

\bibitem{frosst2018distilling}
N.~Frosst et~al.
\newblock Distilling a neural network into a soft decision tree.
\newblock {\em arXiv:1711.09784}, 2017.

\bibitem{goodfellow2014generative}
I.~Goodfellow et~al.
\newblock Generative adversarial nets.
\newblock In {\em NIPS}, 2014.

\bibitem{guidotti2018local}
R.~Guidotti et~al.
\newblock Local rule-based explanations of black box decision systems.
\newblock {\em arXiv:1805.10820}, 2018.

\bibitem{guidotti2019investigating}
R.~Guidotti, A.~Monreale, and L.~Cariaggi.
\newblock Investigating neighborhood generation for explanations of image
  classifiers.
\newblock In {\em PAKDD}, 2019.

\bibitem{guidotti2018survey}
R.~Guidotti, A.~Monreale, S.~Ruggieri, F.~Turini, et~al.
\newblock A survey of methods for explaining black box models.
\newblock {\em ACM CSUR}, 51(5):93:1--42, 2018.

\bibitem{guidotti2019stability}
R.~Guidotti and S.~Ruggieri.
\newblock On the stability of interpretable models. {IJCNN}, 2019.

\bibitem{hara2018maximally}
S.~Hara et~al.
\newblock Maximally invariant data perturbation as explanation.
\newblock {\em arXiv:1806.07004}, 2018.

\bibitem{he2016deep}
K.~He et~al.
\newblock Deep residual learning for image recognition.
\newblock In {\em CVPR}, 2016.

\bibitem{hinton2015distilling}
G.~Hinton et~al.
\newblock Distilling the knowledge in a neural network.
\newblock {\em arXiv:1503.02531}, 2015.

\bibitem{kim2016examples}
B.~Kim et~al.
\newblock Examples are not enough, learn to criticize!
\newblock In {\em NIPS}, 2016.

\bibitem{li2018deep}
O.~Li, H.~Liu, C.~Chen, and C.~Rudin.
\newblock Deep learning for case-based reasoning through prototypes: A neural
  network that explains its predictions. {I}n {AAAI}, 2018.

\bibitem{makhzani2015adversarial}
A.~Makhzani, J.~Shlens, et~al.
\newblock Adversarial autoencoders.
\newblock {\em arXiv:1511.05644}, 2015.

\bibitem{melis2018towards}
D.~A. Melis and T.~Jaakkola.
\newblock Towards robust interpretability with self-explaining neural networks.
\newblock In {\em NIPS}, 2018.

\bibitem{molnar2018interpretable}
C.~Molnar.
\newblock {\em Interpretable machine learning}.
\newblock LeanPub, 2018.

\bibitem{panigutti2019explaining}
C.~Panigutti, R.~Guidotti, A.~Monreale, and D.~Pedreschi.
\newblock Explaining multi-label black-box classifiers for health applications.
\newblock In {\em W3PHIAI}, 2019.

\bibitem{ribeiro2016should}
M.~T. Ribeiro, S.~Singh, and C.~Guestrin.
\newblock Why should i trust you?: Explaining the predictions of any
  classifier.
\newblock In {\em KDD}, pages 1135--1144. ACM, 2016.

\bibitem{shrikumar2016not}
A.~Shrikumar et~al.
\newblock Not just a black box: Learning important features through propagating
  activation differences.
\newblock {\em arXiv:1605.01713}, 2016.

\bibitem{siddharth2016inducing}
N.~Siddharth, B.~Paige, A.~Desmaison, V.~de~Meent, et~al.
\newblock Inducing interpretable representations with variational autoencoders.
\newblock {\em arXiv:1611.07492}, 2016.

\bibitem{simonyan2013deep}
K.~Simonyan, A.~Vedaldi, and A.~Zisserman.
\newblock Deep inside convolutional networks: Visualising image classification
  models and saliency maps.
\newblock {\em arXiv:1312.6034}, 2013.

\bibitem{spinner2018towards}
T.~Spinner et~al.
\newblock Towards an interpretable latent space: an intuitive comparison of
  autoencoders with variational autoencoders.
\newblock In {\em IEEE VIS}, 2018.

\bibitem{sun2019enhancing}
K.~Sun, Z.~Zhu, and Z.~Lin.
\newblock Enhancing the robustness of deep neural networks by boundary
  conditional gan.
\newblock {\em arXiv:1902.11029}, 2019.

\bibitem{sundararajan2017axiomatic}
M.~Sundararajan et~al.
\newblock Axiomatic attribution for dnn.
\newblock In {\em ICML}. JMLR, 2017.

\bibitem{van2018contrastive}
J.~van~der Waa et~al.
\newblock Contrastive explanations with local foil trees.
\newblock {\em arXiv:1806.07470}, 2018.

\bibitem{xie2012image}
J.~Xie et~al.
\newblock Image denoising with deep neural networks.
\newblock In {\em NIPS}, 2012.

\bibitem{zeiler2014visualizing}
M.~D. Zeiler and R.~Fergus.
\newblock Visualizing and understanding convolutional networks.
\newblock In {\em European conference on computer vision}, pages 818--833.
  Springer, 2014.

\end{thebibliography}

\end{document}